\documentclass{article}
\usepackage{natbib}
\usepackage{natbib}
\usepackage{authblk}          % author / affiliation blocks
\usepackage{amsmath,amssymb}  % math symbols
\usepackage{graphicx}         % figures
\usepackage{booktabs}         % professional tables
\usepackage{float}            % figure placement
\usepackage{hyperref}         % hyperlinks
\usepackage{url}              % simple URL typesetting
\usepackage{multirow}
\usepackage{caption}
\usepackage[labelfont=bf]{subcaption}
\usepackage{geometry}         % page margins
\usepackage{xcolor}           % colored text (optional)
\usepackage{enumitem}         % compact lists
\usepackage{microtype}        % better kerning / protrusion
\usepackage{tikz}
\usepackage{parskip}
\usetikzlibrary{shapes.geometric, arrows.meta, positioning}
\tikzset{
  block/.style = {rectangle, draw, thick, minimum width=3cm, minimum height=1cm, align=center, rounded corners},
  arr/.style   = {-{Latex[length=3mm,width=2mm]}, thick}
}

\title{Evaluating Robustness in Latent Diffusion Models\\via Embedding--Level Augmentation}

\author[1]{Boris Martirosyan\thanks{\texttt{bormartirosyan@gmail.com}}}
\author[2]{Aleksei Karmanov\thanks{\texttt{akarmanov@nvidia.com}}}

\affil[1]{American University of Armenia}
\affil[2]{NVIDIA}

\begin{document}
\maketitle

% ---------------------------------------------------------------
\begin{abstract}
Latent diffusion models (LDMs) achieve state-of-the-art performance across various tasks, including image generation and video synthesis. However, they generally lack robustness, a limitation that remains not fully explored in current research. In this paper, we propose several methods to address this gap. First, we hypothesize that the robustness of LDMs primarily should be measured without their text encoder, because if we take and explore the whole architecture, the problems of image generator and text encoders wll be fused. Second, we introduce novel data augmentation techniques designed to reveal robustness shortcomings in LDMs when processing diverse textual prompts. We then fine-tune Stable Diffusion 3 and Stable Diffusion XL models using Dreambooth, incorporating these proposed augmentation methods across multiple tasks. Finally, we propose a novel evaluation pipeline specifically tailored to assess the robustness of LDMs fine-tuned via Dreambooth.
\end{abstract}

% ---------------------------------------------------------------
\section{Introduction}
Latent Diffusion Models (LDMs) \citep{rombach2022ldm}  achieve state-of-the-art performance across various image generation tasks, including high-resolution image synthesis, consistent character generation [3, 4], and many others. During inference, LDMs utilize predefined conditioning information—such as embeddings derived from text prompts, reference images, or other modalities—and latent noise sampled from a Gaussian distribution. This latent noise is iteratively denoised through multiple steps to produce a latent embedding, which subsequently generates a new image after decoding via a pre-trained Variational Autoencoder (VAE) \citep{vae2013}. The resulting images exhibit high realism and visual quality, underscoring the practical applicability of these models in production environments.

\begin{figure}[H]
    \centering
    \includegraphics[width=\textwidth]{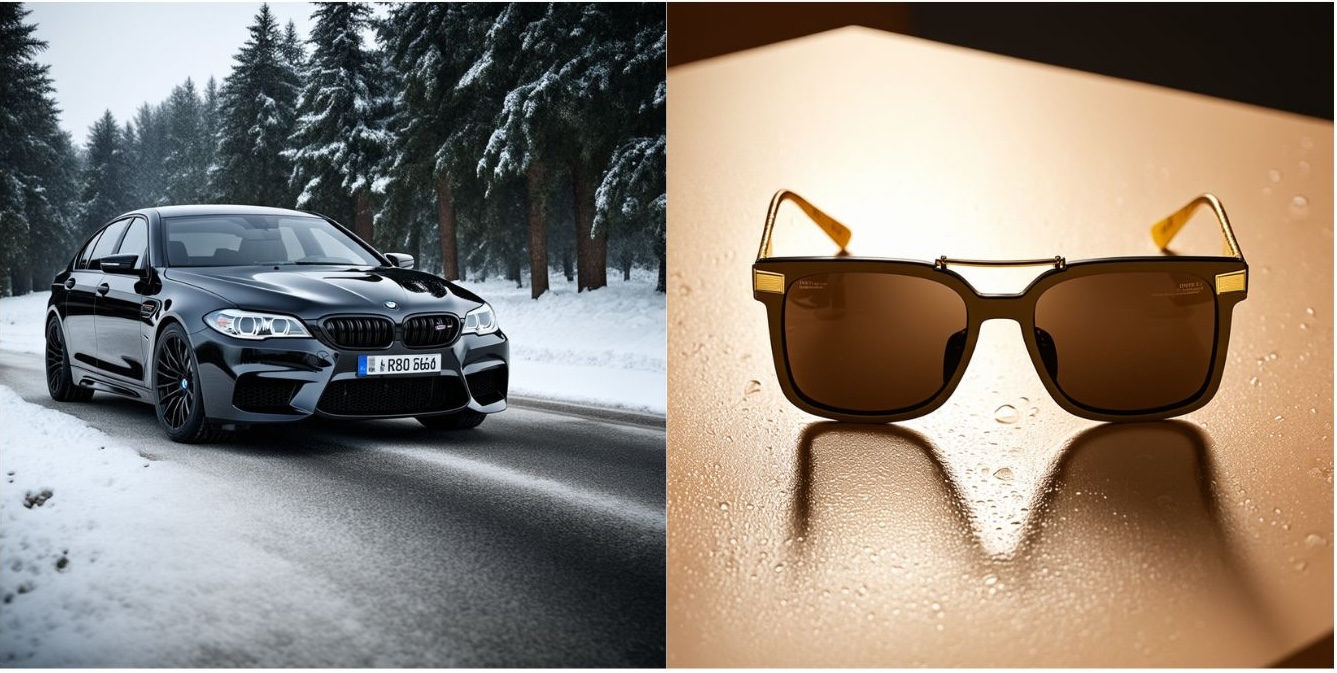} % <-- replace with your image file
    \caption{First (car) image was generated by Stable Diffusion XL, second one (sunglasses) with Stable Diffusion 3. Prompt for the first image is: Create black metallic BMW M5 in a road, front view. Road is covered by snow, around the road are trees. 3D, photorealistic, realistic, great light. Prompt for the second image is: Create an image of sunglasses, with golden tails and rectangle form. Realistic, studio light, on a table. There are water drops on it and around the table.}
\end{figure}

In production and everyday tasks, LDMs are predominantly conditioned on textual inputs, as exemplified by models such as Stable Diffusion \citep{rombach2022ldm, sd2_2022, rombach2023sdxl} and DALLE \citep{ramesh2022dalle2}. Despite their impressive capabilities, these models frequently exhibit a notable lack of robustness in text-to-image generation scenarios. Specifically, our experimentation revealed that generated image quality is highly sensitive to the textual prompts provided; even minor alterations in the input text can result in significantly distorted outputs. Motivated by these observations, we began a systematic investigation into the robustness of LDMs.
Our experimental evaluation focuses primarily on Stable Diffusion 3 (SD3) \citep{watson2024rft} and Stable Diffusion XL (SDXL) \citep{rombach2023sdxl}. To systematically examine the model responses to adversarial inputs, we propose two novel augmentation techniques applied directly to text embeddings: embedding masking (changing token embedding with 0 vector) and embedding convolution with Gaussian-distributed noise. Collectively, we refer to these methods as Augmentation of Embeddings with Latent Implicit Filtering (AELIF).
Importantly, these augmentations are applied after the text encoder(s) and prior to the denoising components of the neural network (such as the Denoising U-net \citep{rombach2022ldm, sd2_2022, rombach2023sdxl} or Diffusion Transformer \citep{watson2024rft}). Applying augmentations before the text encoder would shift the focus away from our primary objective of assessing robustness specifically at the embedding level.
To evaluate our augmentation methods, we selected a diverse set of prompts from the trevordark/sd-prompts \citep{trevordark_sd_prompts} dataset. We then developed a custom inference pipeline enabling visual comparisons between images generated by SDXL and SD3 models, both with and without our proposed augmentations.
Additionally, we introduce a tailored training pipeline for fine-tuning SD3 and SDXL models using Dreambooth \citet{ruiz2022dreambooth}, specifically designed to enhance robustness through exposure to augmented embedding techniques. Finally, we propose two comprehensive evaluation pipelines: the first assesses the overall image quality, while the second specifically evaluates model robustness relative to the original training data.

\noindent Experiments on DreamBooth categories show that AELIF improves robustness while preserving fidelity.

% ---------------------------------------------------------------
\section{Related Work}
Before presenting the core contributions of our research, we provide a brief overview of LDMs and introduce Dreambooth, one of their prominent fine-tuning techniques.
\label{sec:related}
\paragraph{Latent diffusion models.} LDMs were first proposed in the seminal work "High-Resolution Image Synthesis with Latent Diffusion Models" \citet{rombach2022ldm} as a sophisticated advancement over Denoising Diffusion Probabilistic Models (DDPMs) \citep{ho2020ddpm}. The primary training objective of LDMs involves generating high-quality images by progressively denoising Gaussian-distributed latent noise over multiple iterative steps. The training procedure is guided by the following loss function:
\begin{equation}
\mathcal{L}_{LDM} := \mathbb{E}_{\mathcal{E}(x), y, \epsilon \sim \mathcal{N}(0,1), t} \left[ \left\| \epsilon - \epsilon_{\theta}(z_t, t, \tau_{\theta}(y)) \right\|_2^2 \right]
\end{equation}

\begin{itemize}
    \item $\epsilon$ noise added to the latent representation of the training image
    \item $z_t$ image embedding in latent space at timestep $t$
    \item $t$ timestamp during the denoising process
    \item $y$ text prompt
    \item $\tau_{\theta}$ text encoder with parameters $\theta$
    \item $\epsilon_{\theta}$ denoising neural network with parameters $\theta$
    \item $x$ training image
    \item $\mathcal{E}(x)$ encoder of a pretrained variational autoencoder (VAE)
\end{itemize}

\paragraph{DreamBooth.} 
DreamBooth \citet{ruiz2022dreambooth} is  a method for personalizing pre-trained text-to-image diffusion models with user-provided subjects (e.g., specific individuals, objects, or artistic styles), while avoiding catastrophic forgetting of the model’s general capabilities. Built upon the Latent Diffusion framework, DreamBooth fine-tunes both the text encoder and the denoising network using a limited set of instance images that represent the target subject. Simultaneously, it incorporates a larger class dataset to perform prior-preservation regularization.

The training objective extends the standard diffusion noise-prediction loss by introducing a prior-preservation term, which helps mitigate overfitting to the small number of instance images. Specifically, for each timestep t, the loss function is defined as:
\begin{equation}
\mathcal{L}(\theta) = \mathbb{E}_{\epsilon, z_t, t} \left[ \left\| \epsilon - \epsilon_\theta(z_t, t, P_{\text{inst}}) \right\|^2 \right]
+ \lambda \mathbb{E}_{\epsilon, z_t, t} \left[ \left\| \epsilon - \epsilon_\theta(z_t, t, P_{\text{prior}}) \right\|^2 \right]
\end{equation}

where:

\begin{itemize}
    \item $\epsilon$ — Gaussian noise added to the latent embedding
    \item $z_t = \sqrt{\alpha_t} \, \mathcal{E}(x) + \sqrt{1 - \alpha_t} \, \epsilon$ noisy latent at timestep $t$
    \item $\mathcal{E}(\cdot)$ pretrained VAE encoder
    \item $\epsilon_\theta$ U-Net denoising network with parameters $\theta$
    \item $P_{\text{inst}} = [\text{``[V] in a photo!''}]$ tokenized prompt containing the special subject identifier
    \item $P_{\text{prior}} = [\text{``A photo of a C.''}]$ general “class” prompt (e.g., ``A photo of a dog.'')
    \item $\lambda$ scalar weighting the instance-fitting vs prior-preservation terms
\end{itemize}

The DreamBooth training procedure begins by collecting two complementary datasets: an \textit{instance set} of 3--5 photos of the target subject, each paired with a unique token (e.g., ``A photo of [V][V][V] dog in a kitchen''), and a \textit{class set} of approximately 200 generic images of the same class (e.g., ``dog''), each paired with a prompt like ``A photo of a dog.''

Fine-tuning proceeds by initializing the model from pretrained Latent Diffusion checkpoints and, for each mini-batch, sampling half instance pairs $(x_{\text{inst}}, P_{\text{inst}})$ and half class pairs $(x_{\text{prior}}, P_{\text{prior}})$. Each image $x$ is encoded into a latent $z_0 = \mathcal{E}(x)$, and noise is added to produce $z_t$. Both the instance-fitting and prior-preservation noise-prediction losses are then computed.

The prior-preservation term serves as a regularizer, ensuring that the model does not forget general class-level concepts while memorizing the small number of instance images. At inference time, the user can simply prompt ``A photo of [V][V][V] in a park,'' and the model will denoise the corresponding latent through reverse diffusion steps, followed by decoding via the VAE to produce a personalized image.

Through this process, DreamBooth effectively teaches the model to bind a specific identifier token [V][V][V] to the desired concept, while retaining its broader generative capabilities.

\paragraph{Robustness in generative models.} 
Robustness in generative models—such as large language models (LLMs) and diffusion models—refers to their ability to produce accurate, consistent, and reliable outputs across a wide range of inputs, regardless of variations in phrasing, syntax, or contextual nuances. In this work, we aim to improve the robustness of the SDXL and SD3 models. Despite their state-of-the-art performance, our experiments revealed a high sensitivity to adversarial prompts. Even minor grammatical errors or punctuation variations were sufficient to significantly degrade the quality of the generated images.

\begin{figure}[H]
    \centering
    \includegraphics[width=\textwidth]{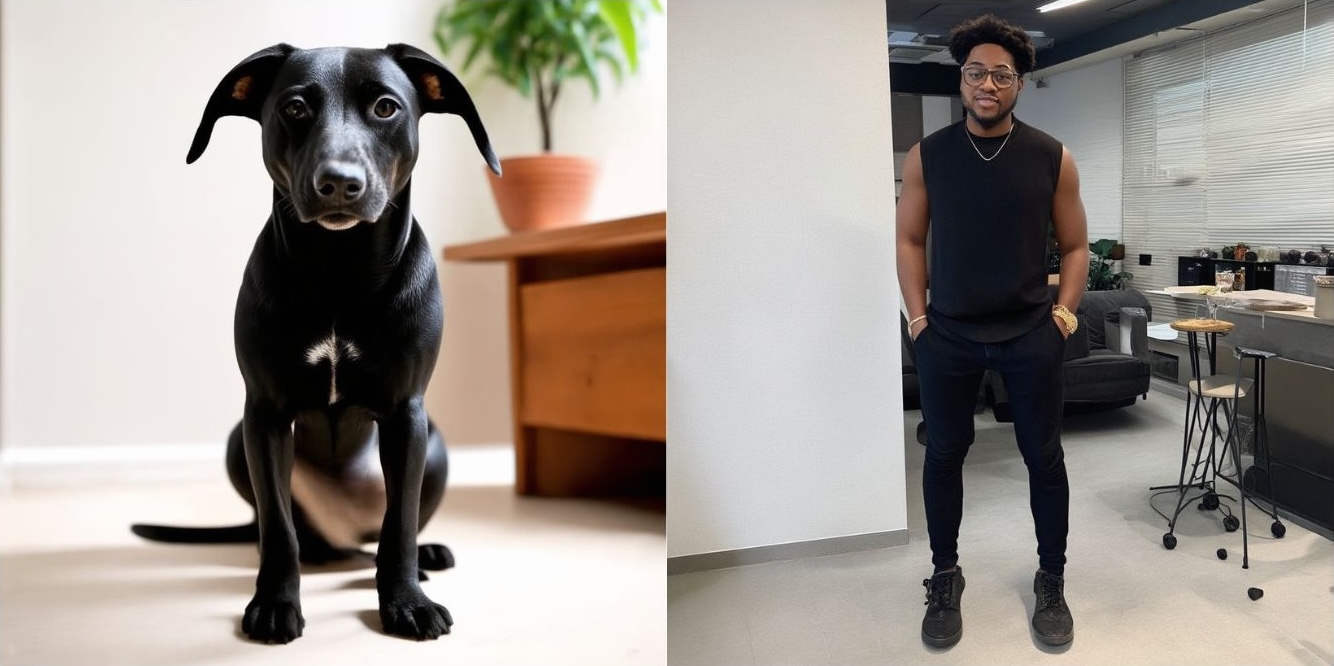} % <-- replace with your image file
    \caption{This figure show casts the problems of robustness in Stable Diffusion 3. For (dog) the first image generation we used this prompt: “An image of dog in black”. For the second (person) image we have used the first prompt’s naturally augmented version. We intentionally removed several letters from the prompt. The second prompt is: “An image of dg in blak”. As you see it resulted in the creation of an image that is very different from the original one.}
\end{figure}

\begin{figure}[H]
    \centering
    \includegraphics[width=\textwidth]{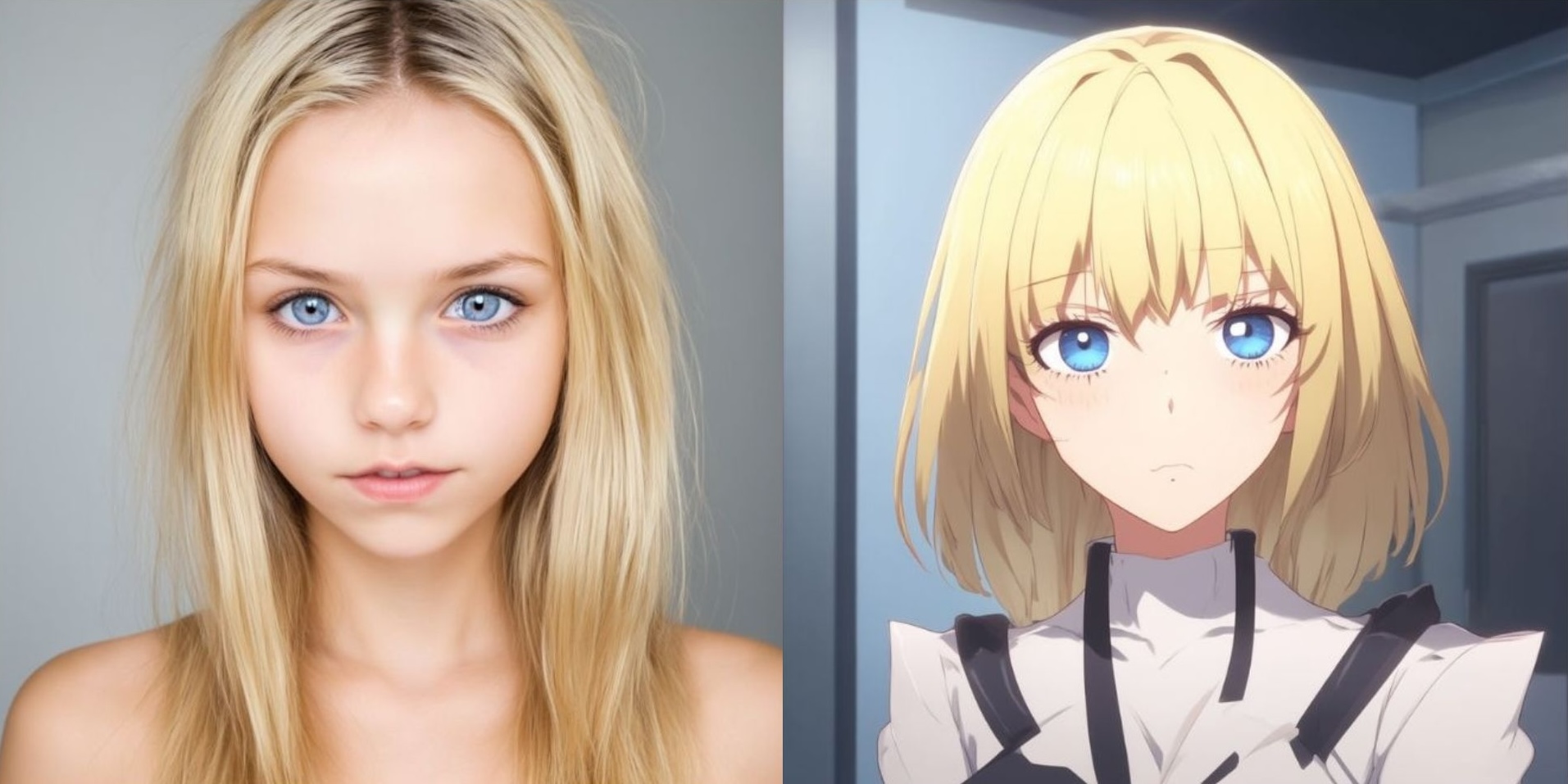} % <-- replace with your image file
    \caption{This figure show casts again a problem with prompt design in Stable Diffusion XL, specifically if we omit several letters in the prompt, the style of generated image can change. For the first image (the photo of a girl with realistic style) we have used this prompt: “An image of a adult girl with blonde hair and blue eyes”. For the second image, we have used first prompt’s augmented version: “Am image of a adlt gil with blonde hair and bue eyes”.}
\end{figure}

% ---------------------------------------------------------------
\section{Methodology}
\subsection{Notation and Preliminaries}

Let $U$ denote the denoising component of the LDM, and let $z_p$ represent the prompt embeddings produced by the text encoder. We define $n$ as a Gaussian noise vector sampled from a standard normal distribution with zero mean and unit variance:
\[
n \sim \mathcal{N}(0, 1)
\]
The function \texttt{aelif\_noise\_conv}($z_p$) applies a noise convolution augmentation to the $z_p$ embeddings, while \texttt{aelif\_mask}($z_p$) performs a masking-based augmentation on the same embeddings.

\subsection{AELIF Augmentation}

We introduce AELIF (Augmentation of Embeddings with Latent Implicit Filtering) and its two core components: \texttt{aelif\_mask} and \texttt{aelif\_noise\_conv}.

\paragraph{aelif\_mask.}
The user specifies an augmentation magnitude $p \in [0, 1]$, representing the fraction of tokens to be masked. Given a sequence of prompt embeddings
\[
Z = \{ z_{p1}, z_{p2}, \dots, z_{pL} \},
\]
where $L$ is the sequence length, the function selects $n = \lfloor L \cdot p \rfloor$ random token positions to be masked. Each selected embedding $z_{pi}$ is then replaced with a zero vector:
\[
z_{pi}' = \mathbf{0}.
\]

\paragraph{aelif\_noise\_conv.}
The user specifies the augmentation magnitude $p \in [0, 1]$, along with a mean and variance for the noise distribution. For each embedding $z_p$ in $Z$, a subset of $n = \lfloor L \cdot p \rfloor$ tokens is selected. For each selected token embedding, a noise vector is sampled from a normal distribution:
\[
\text{noise\_vector} \sim \mathcal{N}(\mu, \sigma^2),
\]
with the same dimensionality as the embeddings. The augmented embedding is then computed via element-wise multiplication:
\[
z_p' = z_p \odot \text{noise\_vector}.
\]

\begin{figure}[H]
    \centering
    \includegraphics[width=\textwidth]{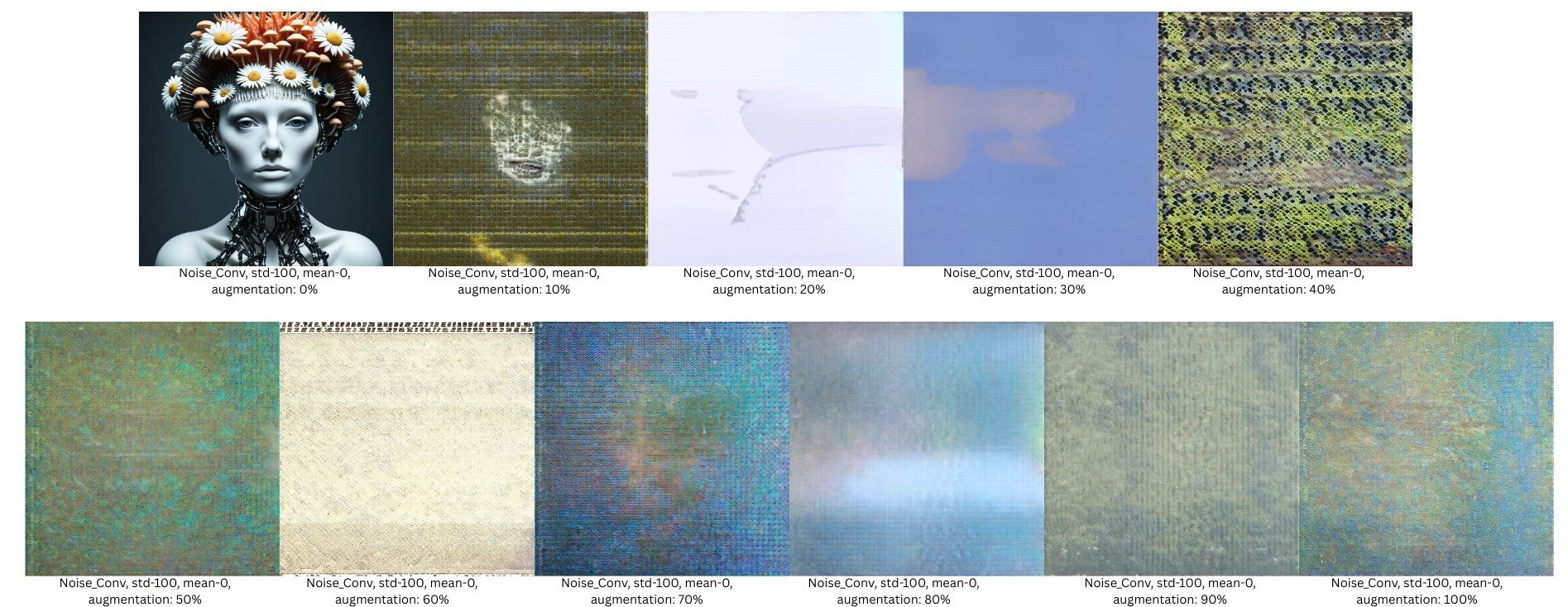} % <-- replace with your image file
    \caption{We took a prompt: “beautiful porcelain ivory fair face woman, close-up, sharp focus, studio light, biomechanical cyborg, iris van herpen haute couture headdress made of rhizomorphs, daisies, brackets, colorful corals, fractal mushrooms, puffballs, octane render, ultra sharp, 8k”. We generated an image with that prompt with Stable Diffusion 3, which you can see in the upper left corner. Then for each iteration we used AELIF noise convolution augmentation, by increasing the augmentation magnitude. The noise mean is taken 0 and standard deviation 100. The figure shows that for even 10 percent of token augmentations, we can have completely distorted output, which is not allowable in this kind of long prompt.}
\end{figure}

\begin{figure}[H]
    \centering
    \includegraphics[width=\textwidth]{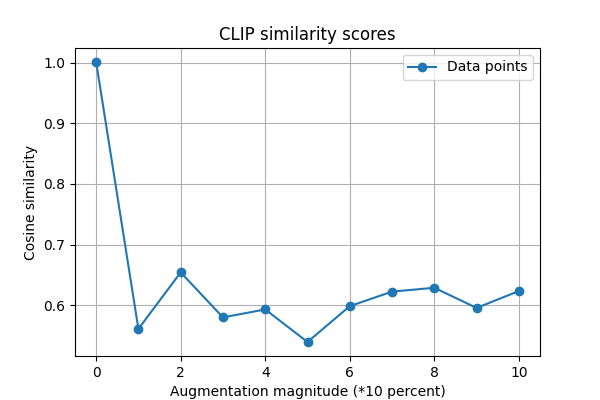} % <-- replace with your image file
    \caption{This figure shows the cosine similarity scores between the CLIP embeddings of the original image (the top-left image in Figure 4) and those of the augmented images. The first point shows the similarity of embeddings of the first image with itself (1 means 100 percent).}
\end{figure}

The intuition behind both augmentation techniques is inspired by the natural errors users often make when interacting with generative models—such as misspellings, grammatical mistakes, and other textual inconsistencies. Masking is the more straightforward of the two, simulating the random omission of characters or tokens, and effectively modeling input degradation due to minor user errors. In contrast, the noise convolution augmentation was designed as a stronger and more comprehensive perturbation method. It aims to capture a wider range of input noise, including grammatical issues, misspellings, and structural distortions such as the random omission or transposition of letters and tokens.

\subsection{Integration into DreamBooth Workflow}

To assess the impact of our proposed augmentations on LDMs, we fine-tuned SDXL and SD3 using the DreamBooth framework.

Due to limited computational resources, we employed DreamBooth in combination with LoRA (Low-Rank Adaptation) \citep{hu2021lora}, a memory-efficient fine-tuning technique that allowed us to adapt large-scale models on commodity hardware.

The DreamBooth dataset provided instance-level supervision, enabling us to establish ground truth references for evaluating the visual fidelity and robustness of generated outputs.

Importantly, the AELIF augmentations were applied \emph{after} the text encoder, directly on the output embeddings. This design isolates the robustness of the denoising architecture without confounding it with potential weaknesses in the text encoder. Applying augmentations before the encoder would entangle the effects, making it difficult to attribute failures or improvements to a specific component of the LDM.

% ---------------------------------------------------------------
\section{Experimental Setup}

\subsection{Training}

For our experiments, we fine-tuned two state-of-the-art latent diffusion models: \textbf{Stable Diffusion~3} and \textbf{Stable Diffusion XL~1.0}. The training dataset consists of several instance categories derived from the public DreamBooth dataset \citep {google2022dreambooth}, available at:
\begin{center}
\url{https://github.com/google/dreambooth}
\end{center}

The following categories (organized by folder name) were used:

\begin{itemize}
    \item \texttt{backpack}
    \item \texttt{candle}
    \item \texttt{dog\_data} (curated externally, see our GitHub repository)
    \item \texttt{cat}
    \item \texttt{colorful\_sneaker}
    \item \texttt{dog2}
    \item \texttt{dog3}
    \item \texttt{backpack\_dog}
    \item \texttt{clock}
    \item \texttt{vase}
    \item \texttt{teapot}
\end{itemize}

For each item, prompts were constructed using the template:
\[
\text{``a photo of sks \textless item\textgreater''}
\]
where \texttt{<item>} corresponds to the cleaned folder name (e.g., \texttt{dog2} is normalized to \texttt{dog}). 

\hfill \break
\hfill \break

\begin{figure}[ht]
  \centering
  \begin{tikzpicture}[node distance=1.5cm and 2cm]
    \node[block] (tokens) {Token Sequence};
    \node[block,right=of tokens] (encoder) {LLM Encoder(s)};
    \node[block,right=of encoder] (aelif) {AELIF Augmentation};
    \node[block,right=of aelif] (denoise) {Denoising\\Neural Network};
    \draw[arr] (tokens) -- (encoder);
    \draw[arr] (encoder) -- (aelif);
    \draw[arr] (aelif) -- (denoise);
  \end{tikzpicture}
  \caption{Overview of our token → encoder → AELIF → denoiser pipeline.}
  \label{fig:pipeline}
\end{figure}
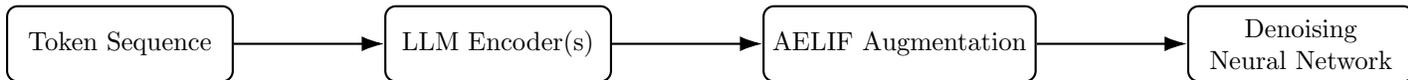

\hfill \break
\hfill \break

\subsection{Evaluation}

We perform two types of evaluation for each checkpoint:

\begin{enumerate}
    \item \textbf{AELIF as a data augmentation technique}
    \item \textbf{AELIF as a robustness enhancement method}
\end{enumerate}

\paragraph{Evaluation Pipeline as Data Augmentation}

\begin{enumerate}[label=\arabic*.]
    \item Train the model \textbf{without} augmentations.
    \item Generate images with fixed seeds (these  seeds count should be equal to training data points count): \\
    $X_O = [x_{o, p_1}, x_{o, p_2}, \dots, x_{o, p_n}]$.
    \item Train the model \textbf{with} AELIF augmentations.
    \item Generate images with the same seeds, but this time with the model, trained with AELIF: \\
    $X_A = [x_{a, p_1}, x_{a, p_2}, \dots, x_{a, p_n}]$.
    \item Compute CLIP embeddings for all outputs and reference:
    \[
    \text{CLIP}(X_O),\quad \text{CLIP}(X_A),\quad \text{CLIP}(X_t)
    \]
    \item For each group of embedding from AELIF augmentation calculate the Wasserstein distance with training training images.
    \[
    W_2(\text{CLIP}(X_A), \text{CLIP}(X_t))
    \]
    \[
    W_2(\text{CLIP}(X_O), \text{CLIP}(X_t))
    \]
    \begin{center}
    lower is better
    \end{center}
    
\end{enumerate}

\paragraph{Robustness Evaluation Pipeline:}

\begin{enumerate}[label=\arabic*.]
    \item Train the model \textbf{without} augmentations.
    \item Generate \textit{adversarial prompts} using the GPT-4o API.
    \item Generate images corresponding to these prompts: \\
    $X_O = [x_{o, p_1}, x_{o, p_2}, \dots, x_{o, p_n}]$.
    \item Train the model \textbf{with} AELIF augmentations.
    \item Generate images for the same adversarial prompts: \\
    $X_A = [x_{a, p_1}, x_{a, p_2}, \dots, x_{a, p_n}]$.
    \item Select a fixed training image $x_t$ for comparison.
    \item Compute CLIP embeddings for all outputs and reference:
    \[
    \text{CLIP}(X_O),\quad \text{CLIP}(X_A),\quad \text{CLIP}(x_t)
    \]
    \item For each image in $X_O$ and $X_A$, compute the 2-Wasserstein distance to $\text{CLIP}(x_t)$:
    \[
    W_2(\text{CLIP}(x), \text{CLIP}(x_t)) \quad \text{(lower is better)}
    \]
\end{enumerate}

\noindent Please refer to the Appendix for evaluation results and comparison tables.

\section{Conclusion and Future Work}

Our experiments demonstrate that embedding-level augmentations—specifically AELIF’s masking and noise convolution strategies—significantly improve the robustness of LDMs without compromising image quality. When integrated into the DreamBooth fine-tuning framework, these augmentations enhance the model's ability to generalize across both clean and adversarial prompt conditions, suggesting their dual role as both regularizers and robustness enablers.

Importantly, the placement of these augmentations after the text encoder isolates their impact on the denoising architecture, providing clearer insight into architectural vulnerabilities and opportunities for improvement.

In future work, we aim to formalize this evaluation into a standardized robustness benchmark for generative models. As diffusion systems are increasingly deployed in real-world applications, stress-testing their behavior under prompt perturbations will be essential. Our planned benchmark will systematically measure performance degradation under noisy, adversarial, and semantically drifted prompts—contributing to a more rigorous and reliable standard for evaluating image generation models at scale.
\newpage

\bibliography{references}   

\begin{thebibliography}{11}
\providecommand{\natexlab}[1]{#1}
\providecommand{\url}[1]{\texttt{#1}}
\expandafter\ifx\csname urlstyle\endcsname\relax
  \providecommand{\doi}[1]{doi: #1}\else
  \providecommand{\doi}{doi: \begingroup \urlstyle{rm}\Url}\fi

\bibitem[AI(2022)]{sd2_2022}
Stability AI.
\newblock Stable diffusion 2.0 and 2.1 release.
\newblock \url{https://stability.ai/blog/stable-diffusion-2-0-release}, 2022.
\newblock Accessed: 2025-05-07.

\bibitem[Dark(2022)]{trevordark_sd_prompts}
Trevor Dark.
\newblock Stable diffusion prompts dataset.
\newblock \url{https://huggingface.co/datasets/trevordark/sd-prompts}, 2022.
\newblock Accessed: 2025-05-07.

\bibitem[Ho et~al.(2020)Ho, Jain, and Abbeel]{ho2020ddpm}
Jonathan Ho, Ajay Jain, and Pieter Abbeel.
\newblock Denoising diffusion probabilistic models.
\newblock \emph{arXiv preprint arXiv:2006.11239}, 2020.
\newblock URL \url{https://arxiv.org/abs/2006.11239}.

\bibitem[Hu et~al.(2021)Hu, Shen, Wallis, Allen-Zhu, Li, Wang, and Chen]{hu2021lora}
Edward~J. Hu, Yelong Shen, Phillip Wallis, Zeyuan Allen-Zhu, Yuanzhi Li, Lu~Wang, and Weizhu Chen.
\newblock Lora: Low-rank adaptation of large language models.
\newblock \emph{arXiv preprint arXiv:2106.09685}, 2021.
\newblock URL \url{https://arxiv.org/abs/2106.09685}.

\bibitem[Kingma and Welling(2013)]{vae2013}
Diederik~P. Kingma and Max Welling.
\newblock Auto-encoding variational bayes.
\newblock \emph{arXiv preprint arXiv:1312.6114}, 2013.
\newblock URL \url{https://arxiv.org/abs/1312.6114}.

\bibitem[Ramesh et~al.(2022)Ramesh, Dhariwal, Nichol, Chu, and Chen]{ramesh2022dalle2}
Aditya Ramesh, Prafulla Dhariwal, Alex Nichol, Casey Chu, and Mark Chen.
\newblock Hierarchical text-conditional image generation with clip latents.
\newblock \emph{arXiv preprint arXiv:2204.06125}, 2022.
\newblock URL \url{https://arxiv.org/abs/2204.06125}.

\bibitem[Research(2022)]{google2022dreambooth}
Google Research.
\newblock Dreambooth dataset.
\newblock \url{https://github.com/google/dreambooth}, 2022.
\newblock Accessed: 2025-05-07.

\bibitem[Rombach et~al.(2021)Rombach, Blattmann, Lorenz, Esser, and Ommer]{rombach2022ldm}
Robin Rombach, Andreas Blattmann, Dominik Lorenz, Patrick Esser, and Björn Ommer.
\newblock High-resolution image synthesis with latent diffusion models.
\newblock \emph{arXiv preprint arXiv:2112.10752}, 2021.
\newblock URL \url{https://arxiv.org/abs/2112.10752}.

\bibitem[Rombach et~al.(2023)Rombach, Tuli, Niculae, Chai, Esser, and Ommer]{rombach2023sdxl}
Robin Rombach, Pradyumna Tuli, Vlad Niculae, Lucy Chai, Patrick Esser, and Björn Ommer.
\newblock Sdxl: Improving latent diffusion models for high-resolution image synthesis.
\newblock \emph{arXiv preprint arXiv:2307.01952}, 2023.
\newblock URL \url{https://arxiv.org/abs/2307.01952}.

\bibitem[Ruiz et~al.(2022)Ruiz, Li, Jozwik, Ginosar, Maschinot, and Zhang]{ruiz2022dreambooth}
Nataniel Ruiz, Sifei Li, Stanislaw Jozwik, Shiry Ginosar, Aaron Maschinot, and Ting Zhang.
\newblock Dreambooth: Fine tuning text-to-image diffusion models for subject-driven generation.
\newblock \emph{arXiv preprint arXiv:2208.12242}, 2022.
\newblock URL \url{https://arxiv.org/abs/2208.12242}.

\bibitem[Watson et~al.(2024)Watson, Karras, Liu, and LeCun]{watson2024rft}
Jack Watson, Tero Karras, Sifei Liu, and Yann LeCun.
\newblock Scaling rectified flow transformers for high-resolution image synthesis.
\newblock \emph{arXiv preprint arXiv:2403.03206}, 2024.
\newblock URL \url{https://arxiv.org/abs/2403.03206}.

\end{thebibliography}

\newpage

\appendix

\section{AELIF Effectiveness via Embedding Distance}

These results demonstrate the effectiveness of AELIF as a data augmentation technique. 
We generated images using three training setups:
\begin{enumerate}
    \item DreamBooth (original)
    \item DreamBooth with AELIF masking
    \item DreamBooth with AELIF noise convolution
\end{enumerate}
We then computed the Wasserstein distance between the CLIP embeddings of the generated images and the training data. Lower distances indicate closer alignment with the training distribution.

\begin{table}[ht]
\centering
\begin{tabular}{lccc}
\hline
\textbf{Item} & \textbf{Noise Conv vs Train} & \textbf{Mask vs Train} & \textbf{Original vs Train} \\
\hline
backpack & 5.19 & 5.19 & 5.21 \\
backpack\_dog & 5.07 & 4.62 & 4.90 \\
candle & 5.10 & 5.54 & 5.18 \\
cat & 3.23 & 2.96 & 3.09 \\
clock & 6.17 & 5.73 & 6.30 \\
colorful\_sneaker & 4.65 & 4.73 & 4.64 \\
dog & 3.95 & 3.95 & 3.98 \\
dog2 & 4.78 & 4.85 & 4.88 \\
dog3 & 5.00 & 5.06 & 4.86 \\
teapot & 4.66 & 5.22 & 4.92 \\
vase & 4.57 & 4.78 & 4.41 \\
\hline
\end{tabular}
\label{tab:sdxl_distances}
\caption{Wasserstein distance between CLIP embeddings of generated images and training data (SDXL model). Lower is better.}
\end{table}

\begin{table}[ht]
\centering

\begin{tabular}{lccc}
\hline
\textbf{Item} & \textbf{Noise Conv vs Train} & \textbf{Mask vs Train} & \textbf{Original vs Train} \\
\hline
backpack & 5.42 & 5.19 & 5.66 \\
backpack\_dog & 8.17 & 7.43 & 7.15 \\
candle & 8.26 & 7.25 & 7.15 \\
cat & 4.08 & 3.84 & 3.83 \\
clock & 5.89 & 5.60 & 5.09 \\
colorful\_sneaker & 5.20 & 5.32 & 5.44 \\
dog & 3.41 & 4.09 & 4.12 \\
dog2 & 4.54 & 4.54 & 4.68 \\
dog3 & 6.66 & 5.96 & 6.20 \\
teapot & 5.70 & 5.43 & 5.47 \\
vase & 5.26 & 4.81 & 5.00 \\
\hline
\end{tabular}
\label{tab:sd3_distances}
\caption{Wasserstein distance between CLIP embeddings of generated images and training data (SD3 model). Lower is better.}
\end{table}

\newpage

\section{Results of Images on Perturbed Prompts}

We collected all prompts generated by the GPT-4o model and performed a robustness evaluation on them, as described earlier. First, we present the summary ("conclusion") of the data frames: for each item category, we show the proportion (in percent) of cases where models trained with our augmentations outperformed those trained without them. Lower is better.

\begin{table}[ht]
\centering
\begin{tabular}{l r}
\hline
\textbf{Category} & \textbf{Proportion (\%)} \\
\hline
backpack & 63.27 \\
backpack\_dog & 52.78 \\
candle & 55.81 \\
cat & 71.43 \\
clock & 73.47 \\
colorful\_sneaker & 68.75 \\
dog2 & 75.51 \\
dog3 & 65.96 \\
dog\_data & 70.97 \\
teapot & 50.00 \\
vase & 66.67 \\
\hline
\textbf{Mean} & \textbf{64.96} \\
\hline
\end{tabular}
\label{tab:sd3_results}
\caption{Robustness evaluation results for SD3 models. Values represent the proportion of cases (in percent) where models trained with our augmentations outperformed those without.}
\end{table}

\begin{table}[ht]
\centering
\begin{tabular}{l r}
\hline
\textbf{Category} & \textbf{Proportion (\%)} \\
\hline
backpack & 70.45 \\
backpack\_dog & 58.33 \\
candle & 72.00 \\
cat & 71.74 \\
clock & 78.38 \\
colorful\_sneaker & 63.64 \\
dog2 & 59.57 \\
dog3 & 62.50 \\
dog\_data & 76.09 \\
teapot & 53.57 \\
vase & 59.52 \\
\hline
\textbf{Mean} & \textbf{65.98} \\
\hline
\end{tabular}
\label{tab:sdxl_results}

\caption{Robustness evaluation results for SDXL models. Values represent the proportion of cases (in percent) where models trained with our augmentations outperformed those without.}
\end{table}

\begin{table}[h]
\centering
\caption{`backpack` category results, SD3}
\label{tab:backpack}
\begin{tabular}{|l|c|c|c|}
\hline
Prompt & Orig & Mask & Noise \\
\hline
a poto of sks backpack & 0.0168 & 0.0201 & 0.0318 \\
a phoo of sks bakcpack & 0.0187 & 0.0202 & 0.0244 \\
a photo of ks backpack & 0.0182 & 0.0189 & 0.0166 \\
a phoot of sks backpakc & 0.0316 & 0.0216 & 0.0188 \\
a phoo of sks bckpack & 0.0198 & 0.0242 & 0.0241 \\
photo of sks backpack & 0.0216 & 0.0207 & 0.0188 \\
a poto of sks bakcpack & 0.0146 & 0.0169 & 0.0288 \\
a phooto of sks backpak & 0.0175 & 0.0159 & 0.0267 \\
a phoo of ks backpack & 0.0250 & 0.0243 & 0.0234 \\
a photo of sks backpck & 0.0203 & 0.0166 & 0.0209 \\
a phto of sks bakcpack & 0.0204 & 0.0177 & 0.0335 \\
a phoot of ks bacpack & 0.0172 & 0.0176 & 0.0243 \\
a photo of sks bckpack & 0.0246 & 0.0221 & 0.0201 \\
a pptoo of sks bckpack & 0.0173 & 0.0231 & 0.0233 \\
a poto of sks bacpak & 0.0238 & 0.0174 & 0.0360 \\
a photo of ss backpack & 0.0269 & 0.0237 & 0.0302 \\
a poto o sks backpack & 0.0151 & 0.0159 & 0.0229 \\
a phpto of ks backpack & 0.0203 & 0.0166 & 0.0337 \\
a phoo of ssk backpack & 0.0205 & 0.0140 & 0.0191 \\
a photo of sks bacpack & 0.0194 & 0.0216 & 0.0241 \\
a poto of sks ckpack & 0.0259 & 0.0208 & 0.0346 \\
a photo of sks bckpakc & 0.0323 & 0.0270 & 0.0272 \\
a phoo of sks pacpack & 0.0213 & 0.0507 & 0.0384 \\
a phoo of sks bkpack & 0.0203 & 0.0209 & 0.0186 \\
photo of sks bakaack & 0.0314 & 0.0302 & 0.0321 \\
a photo f sks backpack & 0.0197 & 0.0195 & 0.0228 \\
aphoto of sks bakcpack & 0.0176 & 0.0178 & 0.0230 \\
a phoo of ks bacpack & 0.0160 & 0.0190 & 0.0194 \\
a poto of sks bckpack & 0.0277 & 0.0251 & 0.0304 \\
a phhot of ks backpack & 0.0205 & 0.0251 & 0.0331 \\
photo of sks bckpak & 0.0220 & 0.0268 & 0.0362 \\
a phto of ssk backpack & 0.0224 & 0.0168 & 0.0219 \\
a photto of kk backpack & 0.0336 & 0.0263 & 0.0224 \\
a photo of sks backpak & 0.0205 & 0.0233 & 0.0151 \\
a photo of sks bakpack & 0.0236 & 0.0250 & 0.0236 \\
a poto o ssk backpack & 0.0189 & 0.0153 & 0.0207 \\
a poho of ks backpack & 0.0181 & 0.0179 & 0.0268 \\
apoto of sks backpakc & 0.0309 & 0.0400 & 0.0238 \\
a poto of ssk bacpack & 0.0157 & 0.0195 & 0.0193 \\
a phoo of ssk bacpack & 0.0224 & 0.0229 & 0.0172 \\
a pohto of sksocbackpack & 0.0213 & 0.0263 & 0.0200 \\
a phoof ss backpack & 0.0266 & 0.0238 & 0.0219 \\
a phoo of ssk ackpack & 0.0229 & 0.0211 & 0.0294 \\
a photo of sks bacpak & 0.0191 & 0.0205 & 0.0222 \\
a photo o sks bacpack & 0.0172 & 0.0174 & 0.0253 \\
a phoo of ks bakcpack & 0.0200 & 0.0182 & 0.0224 \\
a photo of sks bacpac & 0.0228 & 0.0164 & 0.0234 \\
a poto of sks bcpack & 0.0205 & 0.0249 & 0.0411 \\
a phto of sk sistem backpack & 0.0211 & 0.0225 & 0.0189 \\
\hline
\end{tabular}
\end{table}
%--------------------------------------------------------------------------------
\begin{table}[h]
\centering
\caption{`backpack dog` category results, SD3}
\label{tab:backpack_dog}
\begin{tabular}{|l|c|c|c|}
\hline
Prompt & Orig & Mask & Noise \\
\hline
a phoo of sks backpack & 0.0261 & 0.0322 & 0.0232 \\
a poto of sks backpak & 0.0284 & 0.0254 & 0.0253 \\
a phto of sks backpak & 0.0218 & 0.0212 & 0.0285 \\
a phoo of sks bacpack & 0.0232 & 0.0264 & 0.0281 \\
a photo of sks bakpack & 0.0244 & 0.0339 & 0.0372 \\
a pohot of sks bacpack & 0.0283 & 0.0223 & 0.0259 \\
a phott of sks backpak & 0.0209 & 0.0273 & 0.0310 \\
a poto of sks backpack & 0.0190 & 0.0267 & 0.0259 \\
a phoo of sks bakcpack & 0.0210 & 0.0282 & 0.0342 \\
a poto of sks bakpack & 0.0254 & 0.0209 & 0.0263 \\
a phoot of sks bacpak & 0.0312 & 0.0230 & 0.0258 \\
a phot of sks bakcpack & 0.0249 & 0.0331 & 0.0293 \\
a phott of sks bakpack & 0.0217 & 0.0193 & 0.0280 \\
a poto of sks bacpack & 0.0239 & 0.0252 & 0.0283 \\
a photo of sks backpak & 0.0176 & 0.0309 & 0.0370 \\
a poto of sks bakcpack & 0.0222 & 0.0250 & 0.0301 \\
a phto of sks bakcpack & 0.0203 & 0.0221 & 0.0274 \\
a hoto of sks bakpack & 0.0223 & 0.0285 & 0.0276 \\
a poto of sks bakpak & 0.0258 & 0.0218 & 0.0221 \\
a phto of sks bacpack & 0.0220 & 0.0213 & 0.0231 \\
a phoo of sks bkpack & 0.0233 & 0.0311 & 0.0363 \\
a photo of sks pakpack & 0.0201 & 0.0215 & 0.0309 \\
a otho of sks backpak & 0.0247 & 0.0215 & 0.0263 \\
a poto of sks bacpak & 0.0286 & 0.0277 & 0.0278 \\
a phoo of sks bckpack & 0.0260 & 0.0237 & 0.0302 \\
a oto of sks bakpack & 0.0238 & 0.0214 & 0.0259 \\
a poto of sks bckpack & 0.0186 & 0.0237 & 0.0299 \\
a otho of sks ckpack & 0.0378 & 0.0286 & 0.0346 \\
a ohto of sks bakpack & 0.0244 & 0.0200 & 0.0316 \\
a oto of sks ackpack & 0.0256 & 0.0257 & 0.0252 \\
a phto of sks bakpak & 0.0239 & 0.0210 & 0.0226 \\
a photo of sks bakpak & 0.0244 & 0.0301 & 0.0333 \\
a hoto of sks backpak & 0.0211 & 0.0250 & 0.0286 \\
a phott of sks bacpak & 0.0252 & 0.0225 & 0.0334 \\
a photo of sks bcpack & 0.0235 & 0.0322 & 0.0292 \\
a photo of sks bckpack & 0.0258 & 0.0248 & 0.0334 \\
\hline
\end{tabular}
\end{table}
%--------------------------------------------------------------------------------
\begin{table}[h]
\centering
\caption{`candle` category results, SD3}
\label{tab:candle}
\begin{tabular}{|l|c|c|c|}
\hline
Prompt & Orig & Mask & Noise \\
\hline
a poto of sks candle & 0.0143 & 0.0277 & 0.0169 \\
a phoo of sks cadle & 0.0209 & 0.0248 & 0.0238 \\
a phoot of sks cndle & 0.0266 & 0.0235 & 0.0246 \\
a phto of sks candel & 0.0253 & 0.0252 & 0.0244 \\
a phoo of sks cndel & 0.0218 & 0.0216 & 0.0226 \\
a phoo of sks candel & 0.0218 & 0.0236 & 0.0231 \\
a phote of sks cndle & 0.0222 & 0.0226 & 0.0240 \\
a phto of sks cadle & 0.0276 & 0.0243 & 0.0235 \\
a poto of sks candel & 0.0221 & 0.0242 & 0.0227 \\
a phto of sks cndl & 0.0259 & 0.0196 & 0.0231 \\
a phto of sks cndle & 0.0255 & 0.0210 & 0.0229 \\
a poto of sks cndel & 0.0275 & 0.0210 & 0.0233 \\
a poto of sks cndle & 0.0220 & 0.0255 & 0.0244 \\
a photo of sks candl & 0.0279 & 0.0233 & 0.0274 \\
a phoot of sks canle & 0.0255 & 0.0236 & 0.0282 \\
a photo of sks cande & 0.0318 & 0.0334 & 0.0280 \\
a phoo of sks candl & 0.0229 & 0.0242 & 0.0212 \\
a phoo of sks cndle & 0.0212 & 0.0219 & 0.0219 \\
a photo of sks cndel & 0.0219 & 0.0212 & 0.0277 \\
a phto of sks cndel & 0.0262 & 0.0286 & 0.0222 \\
a phoo of sks canle & 0.0232 & 0.0230 & 0.0281 \\
a phote of sks candel & 0.0221 & 0.0231 & 0.0282 \\
a poto of sks canle & 0.0294 & 0.0221 & 0.0228 \\
a poto of sks candl & 0.0214 & 0.0250 & 0.0218 \\
a photo of sks cndl & 0.0193 & 0.0210 & 0.0259 \\
a phoot of sks cndl & 0.0211 & 0.0242 & 0.0243 \\
a phote of sks cndel & 0.0290 & 0.0218 & 0.0228 \\
a photo of sks cadle & 0.0208 & 0.0217 & 0.0306 \\
a phote of sks candl & 0.0242 & 0.0218 & 0.0277 \\
a poto of sks canel & 0.0204 & 0.0221 & 0.0226 \\
a phote of sks candle & 0.0169 & 0.0293 & 0.0278 \\
a photo of sks cndle & 0.0232 & 0.0229 & 0.0283 \\
a phote of sks canle & 0.0207 & 0.0248 & 0.0248 \\
a phote of sks cndl & 0.0206 & 0.0201 & 0.0253 \\
a poth of sks cndle & 0.0268 & 0.0244 & 0.0273 \\
a photo of sks canel & 0.0273 & 0.0212 & 0.0258 \\
a poto of sks canld & 0.0212 & 0.0194 & 0.0237 \\
a poto of sks cndl & 0.0216 & 0.0214 & 0.0243 \\
a poto of sks canlde & 0.0218 & 0.0238 & 0.0251 \\
a poho of sks cndle & 0.0227 & 0.0258 & 0.0230 \\
a poto of sks cnadle & 0.0222 & 0.0242 & 0.0249 \\
a photo of sks candel & 0.0215 & 0.0206 & 0.0181 \\
a potho of sks cndle & 0.0202 & 0.0270 & 0.0332 \\
\hline
\end{tabular}
\end{table}
%--------------------------------------------------------------------------------
\begin{table}[h]
\centering
\caption{`cat` category results, SD3}
\label{tab:cat}
\begin{tabular}{|l|c|c|c|}
\hline
Prompt & Orig & Mask & Noise \\
\hline
a poto of sks cat & 0.0185 & 0.0112 & 0.0233 \\
a phoot of sks cat & 0.0171 & 0.0156 & 0.0231 \\
a ppoto of sks cat & 0.0174 & 0.0118 & 0.0202 \\
a phto of sks cat & 0.0163 & 0.0139 & 0.0215 \\
a phot of sks cat & 0.0163 & 0.0123 & 0.0187 \\
a phoo of sks cat & 0.0186 & 0.0157 & 0.0163 \\
a pohot of sks cat & 0.0168 & 0.0157 & 0.0222 \\
a photo o sks cat & 0.0164 & 0.0177 & 0.0166 \\
a pohoto of sks cat & 0.0157 & 0.0171 & 0.0213 \\
a poto f sks cat & 0.0158 & 0.0145 & 0.0214 \\
a photo  of sks cat & 0.0199 & 0.0137 & 0.0153 \\
a phot o sks cat & 0.0135 & 0.0129 & 0.0190 \\
a photo  sks cat & 0.0146 & 0.0147 & 0.0211 \\
a photo of ks cat & 0.0147 & 0.0163 & 0.0201 \\
\hline
\end{tabular}
\end{table}
%--------------------------------------------------------------------------------
\begin{table}[h]
\centering
\caption{`clock` category results, SD3}
\label{tab:clock}
\begin{tabular}{|l|c|c|c|}
\hline
Prompt & Orig & Mask & Noise \\
\hline
a poto of sks clok & 0.0295 & 0.0260 & 0.0321 \\
a phoo of sks clock & 0.0288 & 0.0275 & 0.0229 \\
a photo f sks cloc & 0.0327 & 0.0293 & 0.0305 \\
a phoot of sks clok & 0.0347 & 0.0331 & 0.0326 \\
a pohoto of sks cloock & 0.0269 & 0.0271 & 0.0325 \\
a poto of sks clcok & 0.0298 & 0.0251 & 0.0346 \\
a phoo f sks clk & 0.0358 & 0.0374 & 0.0379 \\
a phoot of sks clcok & 0.0284 & 0.0293 & 0.0363 \\
a ohoto of sks clok & 0.0305 & 0.0272 & 0.0255 \\
a phto of sks clock & 0.0303 & 0.0226 & 0.0216 \\
a phoo of sks cloock & 0.0336 & 0.0282 & 0.0313 \\
a phooof sks clok & 0.0291 & 0.0249 & 0.0350 \\
a potoo of sks clok & 0.0386 & 0.0354 & 0.0322 \\
a photto of sks cloock & 0.0279 & 0.0286 & 0.0297 \\
a phto of sks clcok & 0.0305 & 0.0251 & 0.0329 \\
a pote of sks clock & 0.0263 & 0.0327 & 0.0188 \\
a phoot of sks clk & 0.0310 & 0.0297 & 0.0323 \\
a pohot of sks clok & 0.0265 & 0.0297 & 0.0317 \\
a phoo of sks clcok & 0.0301 & 0.0256 & 0.0364 \\
a phoco of sks clok & 0.0315 & 0.0238 & 0.0292 \\
a phoo of sks cllok & 0.0286 & 0.0292 & 0.0301 \\
a oto of sks cloc & 0.0276 & 0.0218 & 0.0316 \\
a phohto of sks cloock & 0.0285 & 0.0277 & 0.0332 \\
a phot of sks clck & 0.0295 & 0.0276 & 0.0299 \\
a pho of sks clk & 0.0304 & 0.0300 & 0.0322 \\
a oo of sks clok & 0.0293 & 0.0215 & 0.0307 \\
a phtoto of sks clok & 0.0247 & 0.0286 & 0.0309 \\
a phto of sks cloock & 0.0333 & 0.0282 & 0.0329 \\
a phooo of sks clok & 0.0255 & 0.0242 & 0.0317 \\
a poto of sks cclock & 0.0222 & 0.0232 & 0.0287 \\
a pohot of sks cloc & 0.0292 & 0.0261 & 0.0348 \\
a phoo of sks cllcok & 0.0384 & 0.0407 & 0.0356 \\
a phtoo of sks clck & 0.0272 & 0.0278 & 0.0360 \\
a phpt of sks clok & 0.0289 & 0.0253 & 0.0310 \\
a phpot of sks clok & 0.0205 & 0.0265 & 0.0261 \\
a ohhoo of sks cloc & 0.0240 & 0.0303 & 0.0287 \\
a poto of sks cllok & 0.0293 & 0.0280 & 0.0304 \\
a opo of sks clock & 0.0260 & 0.0250 & 0.0226 \\
a poto of sks cllk & 0.0286 & 0.0263 & 0.0334 \\
a pho of sks clcok & 0.0336 & 0.0369 & 0.0331 \\
a phyyy of sks clck & 0.0269 & 0.0255 & 0.0345 \\
a phooto of sks cloc & 0.0334 & 0.0237 & 0.0321 \\
a pof of sks clok & 0.0277 & 0.0316 & 0.0275 \\
a fhoo of sks clok & 0.0222 & 0.0256 & 0.0352 \\
a pht of sks clok & 0.0317 & 0.0224 & 0.0265 \\
a pot of sks clock & 0.0393 & 0.0313 & 0.0167 \\
a photo of sks cllk & 0.0265 & 0.0337 & 0.0279 \\
a poto of sks cllcok & 0.0287 & 0.0263 & 0.0434 \\
a phots of sks clock & 0.0259 & 0.0326 & 0.0250 \\
\hline
\end{tabular}
\end{table}
%--------------------------------------------------------------------------------

\begin{table}[h]
\centering
\caption{`colorful sneaker` category results, SD3}
\label{tab:colorful_sneaker}
\begin{tabular}{|l|c|c|c|}
\hline
Prompt & Orig & Mask & Noise \\
\hline
a poto of sks sneaker & 0.0202 & 0.0214 & 0.0176 \\
a phot of sks sneaker & 0.0215 & 0.0252 & 0.0243 \\
a phto of sks snaker & 0.0384 & 0.0326 & 0.0319 \\
a phoot of sks sneaker & 0.0209 & 0.0233 & 0.0249 \\
a poto of sks sneker & 0.0330 & 0.0322 & 0.0266 \\
a phto of sks sneker & 0.0351 & 0.0263 & 0.0292 \\
a phoot of sks sneker & 0.0330 & 0.0225 & 0.0275 \\
a potof sks sneaker & 0.0266 & 0.0294 & 0.0177 \\
a photof sks snker & 0.0337 & 0.0434 & 0.0387 \\
a phto of sks sneaker & 0.0215 & 0.0227 & 0.0210 \\
a photo of sks sneker & 0.0295 & 0.0310 & 0.0351 \\
a phot of sks snker & 0.0318 & 0.0303 & 0.0319 \\
a poto of sks snker & 0.0304 & 0.0285 & 0.0292 \\
a potof sks sneker & 0.0305 & 0.0333 & 0.0327 \\
a phon of sks sneker & 0.0291 & 0.0388 & 0.0382 \\
a poto o sks sneaker & 0.0266 & 0.0209 & 0.0251 \\
a phoot of sks snker & 0.0355 & 0.0312 & 0.0347 \\
a pho of sks sneaker & 0.0266 & 0.0234 & 0.0210 \\
a poto of sks snaker & 0.0355 & 0.0312 & 0.0368 \\
a pht of sks sneker & 0.0410 & 0.0334 & 0.0364 \\
a poto of sks snnr & 0.0272 & 0.0332 & 0.0277 \\
a phoo of sks snaker & 0.0280 & 0.0303 & 0.0395 \\
a phoo of sks snker & 0.0354 & 0.0346 & 0.0393 \\
a poto o sks sneker & 0.0294 & 0.0268 & 0.0264 \\
a phto f sks sneaker & 0.0225 & 0.0250 & 0.0253 \\
a poto of sks snek & 0.0339 & 0.0321 & 0.0256 \\
a phoo of sks sner & 0.0418 & 0.0266 & 0.0346 \\
a phoo of sks sneker & 0.0406 & 0.0364 & 0.0278 \\
a poto of sks snekoer & 0.0251 & 0.0365 & 0.0341 \\
a phoot of sks snek & 0.0443 & 0.0293 & 0.0374 \\
a phoo of sks snr & 0.0362 & 0.0324 & 0.0347 \\
a poto of sks sner & 0.0257 & 0.0315 & 0.0300 \\
a pho of sks snaker & 0.0353 & 0.0356 & 0.0354 \\
a poht of sks sneaker & 0.0246 & 0.0243 & 0.0246 \\
a pohto of sks sneaker & 0.0224 & 0.0215 & 0.0205 \\
a poto of sks sneake & 0.0311 & 0.0337 & 0.0350 \\
a phto of sks snake & 0.0390 & 0.0324 & 0.0322 \\
a poto of sks skneaker & 0.0373 & 0.0324 & 0.0343 \\
a poto fo sks sneaker & 0.0229 & 0.0222 & 0.0240 \\
a phoo of sks nnker & 0.0357 & 0.0371 & 0.0298 \\
a poto of sks snekae & 0.0342 & 0.0322 & 0.0309 \\
a poto fo sks snker & 0.0364 & 0.0369 & 0.0353 \\
a poto fo sks sneker & 0.0383 & 0.0321 & 0.0307 \\
a phto o sks sneaker & 0.0228 & 0.0271 & 0.0205 \\
a poto f sks snker & 0.0260 & 0.0420 & 0.0265 \\
a poto of sks skeaker & 0.0342 & 0.0365 & 0.0256 \\
a poto of sks sneakr & 0.0359 & 0.0352 & 0.0311 \\
a poto of sks sneaer & 0.0337 & 0.0350 & 0.0371 \\
\hline
\end{tabular}
\end{table}
%--------------------------------------------------------------------------------
\begin{table}[h]
\centering
\caption{`dog2` category results, SD3}
\label{tab:dog2}
\begin{tabular}{|l|c|c|c|}
\hline
Prompt & Orig & Mask & Noise \\
\hline
a phto of sks dog & 0.0198 & 0.0180 & 0.0221 \\
a foto of sks dog & 0.0157 & 0.0171 & 0.0183 \\
a poto of sks dog & 0.0164 & 0.0192 & 0.0219 \\
a phoo of sks dg & 0.0322 & 0.0301 & 0.0305 \\
a phot of sks do & 0.0313 & 0.0287 & 0.0302 \\
a phoo of sks dog & 0.0150 & 0.0199 & 0.0181 \\
a phto of sks dg & 0.0349 & 0.0377 & 0.0243 \\
a photo of ssk dog & 0.0188 & 0.0172 & 0.0180 \\
a poto of sks dg & 0.0287 & 0.0378 & 0.0236 \\
a phoo of ssk dog & 0.0183 & 0.0183 & 0.0238 \\
a foro of sks dog & 0.0183 & 0.0178 & 0.0187 \\
a foto of sks dg & 0.0369 & 0.0417 & 0.0231 \\
a poto of ssk dog & 0.0167 & 0.0208 & 0.0234 \\
a phoot of sks dog & 0.0210 & 0.0179 & 0.0223 \\
a phto of sks do & 0.0272 & 0.0321 & 0.0318 \\
a poto of ssk dg & 0.0257 & 0.0281 & 0.0222 \\
a photo of sks dg & 0.0414 & 0.0438 & 0.0156 \\
a phot of sks dg & 0.0314 & 0.0317 & 0.0238 \\
a foro of sks dg & 0.0251 & 0.0311 & 0.0194 \\
a potoof sks dg & 0.0282 & 0.0315 & 0.0242 \\
a phto of scs dog & 0.0171 & 0.0173 & 0.0214 \\
a foro of ssk dog & 0.0211 & 0.0239 & 0.0233 \\
a potoo of sks dog & 0.0239 & 0.0293 & 0.0199 \\
a photo of ska dog & 0.0176 & 0.0162 & 0.0185 \\
a phoo of ssk dg & 0.0418 & 0.0373 & 0.0229 \\
a poto of ska dog & 0.0188 & 0.0186 & 0.0167 \\
a fot of sks dog & 0.0210 & 0.0191 & 0.0219 \\
a photoo of sks dog & 0.0153 & 0.0190 & 0.0203 \\
a phto of ssk do & 0.0333 & 0.0317 & 0.0211 \\
a phott of sks dog & 0.0194 & 0.0202 & 0.0172 \\
a poto of skk dog & 0.0210 & 0.0199 & 0.0203 \\
a photo of sks do & 0.0355 & 0.0274 & 0.0198 \\
a phoot of sks dg & 0.0344 & 0.0322 & 0.0345 \\
a phoo of ska dog & 0.0189 & 0.0217 & 0.0183 \\
a poto of ska dg & 0.0346 & 0.0320 & 0.0198 \\
a ptoto of sks dog & 0.0169 & 0.0217 & 0.0196 \\
a phocto of sks dog & 0.0168 & 0.0186 & 0.0183 \\
a poto of ssk do & 0.0217 & 0.0222 & 0.0188 \\
a fhto of sks dog & 0.0164 & 0.0161 & 0.0166 \\
a phot of ssk dog & 0.0167 & 0.0223 & 0.0168 \\
a poto of skk dg & 0.0424 & 0.0308 & 0.0361 \\
a phoo of skk dog & 0.0179 & 0.0192 & 0.0168 \\
a ptoto of ssk dog & 0.0184 & 0.0176 & 0.0171 \\
a phoo of ski dog & 0.0265 & 0.0267 & 0.0258 \\
a photo of skz dog & 0.0203 & 0.0164 & 0.0180 \\
a photo os sks dog & 0.0180 & 0.0176 & 0.0168 \\
a pho of sks dog & 0.0358 & 0.0298 & 0.0391 \\
a photo of sks fog & 0.0299 & 0.0252 & 0.0264 \\
a phoo of sks oda & 0.0271 & 0.0258 & 0.0185 \\
\hline
\end{tabular}
\end{table}
%--------------------------------------------------------------------------------
\begin{table}[h]
\centering
\caption{`dog3` category results, SD3}
\label{tab:dog3}
\begin{tabular}{|l|c|c|c|}
\hline
Prompt & Orig & Mask & Noise \\
\hline
a poto of sks dog & 0.0222 & 0.0210 & 0.0178 \\
a phoo of sks do & 0.0257 & 0.0286 & 0.0279 \\
a phoo of sks dg & 0.0295 & 0.0260 & 0.0295 \\
a phto of sks dog & 0.0195 & 0.0222 & 0.0314 \\
a pohto of sks dog & 0.0269 & 0.0231 & 0.0234 \\
a phto of sks dg & 0.0293 & 0.0499 & 0.0379 \\
a poto of sks dg & 0.0277 & 0.0268 & 0.0269 \\
a pot of sks dog & 0.0245 & 0.0257 & 0.0188 \\
a ohto of sks dog & 0.0316 & 0.0226 & 0.0226 \\
a phoo of sks dgo & 0.0329 & 0.0336 & 0.0272 \\
a poto of sks dgo & 0.0254 & 0.0281 & 0.0311 \\
a photo o sks dog & 0.0219 & 0.0218 & 0.0209 \\
a potho of sks dog & 0.0188 & 0.0202 & 0.0260 \\
a poto of ks dog & 0.0272 & 0.0201 & 0.0235 \\
a pho of sks dog & 0.0240 & 0.0247 & 0.0244 \\
a pote of sks dog & 0.0238 & 0.0223 & 0.0192 \\
a pho of sks dg & 0.0234 & 0.0243 & 0.0246 \\
a phot of sks do & 0.0247 & 0.0261 & 0.0278 \\
a poo of sks dog & 0.0280 & 0.0233 & 0.0217 \\
a phot of sks dg & 0.0235 & 0.0295 & 0.0278 \\
a phoho of sks dog & 0.0332 & 0.0198 & 0.0207 \\
a photo fo sks dog & 0.0211 & 0.0258 & 0.0204 \\
a pho of sks do & 0.0316 & 0.0269 & 0.0252 \\
a phoof sks dog & 0.0210 & 0.0203 & 0.0170 \\
a phoo f sks dog & 0.0200 & 0.0223 & 0.0192 \\
a phoot of sks dog & 0.0238 & 0.0231 & 0.0237 \\
a photo of ks dog & 0.0251 & 0.0250 & 0.0257 \\
a poto of kks dog & 0.0247 & 0.0304 & 0.0257 \\
a poto of sks do & 0.0295 & 0.0272 & 0.0305 \\
a photo of sks dg & 0.0238 & 0.0276 & 0.0248 \\
a poto of sks og & 0.0258 & 0.0306 & 0.0268 \\
a pohto f sks dog & 0.0239 & 0.0197 & 0.0281 \\
a potoo of sks dog & 0.0406 & 0.0462 & 0.0440 \\
a hoto of sks dog & 0.0197 & 0.0220 & 0.0182 \\
a phto fo sks dog & 0.0231 & 0.0183 & 0.0260 \\
a phto of sk dog & 0.0248 & 0.0190 & 0.0159 \\
a pohot of sks dog & 0.0171 & 0.0201 & 0.0190 \\
a photo fo sks dg & 0.0397 & 0.0393 & 0.0260 \\
a photo o sks dg & 0.0433 & 0.0418 & 0.0173 \\
a poto of kks dg & 0.0423 & 0.0428 & 0.0347 \\
a poto of skss dog & 0.0236 & 0.0267 & 0.0202 \\
a phoo of skss dg & 0.0367 & 0.0406 & 0.0483 \\
a poto of kss dog & 0.0272 & 0.0244 & 0.0203 \\
a phoo fo sks dug & 0.0195 & 0.0252 & 0.0255 \\
a phto of sks do & 0.0253 & 0.0290 & 0.0233 \\
a photo of ss dog & 0.0232 & 0.0208 & 0.0231 \\
a photo of ks do & 0.0346 & 0.0402 & 0.0296 \\
\hline
\end{tabular}
\end{table}
%--------------------------------------------------------------------------------
\begin{table}[h]
\centering
\caption{`dog data` category results, SD3}
\label{tab:dog_data}
\begin{tabular}{|l|c|c|c|}
\hline
Prompt & Orig & Mask & Noise \\
\hline
a poto of sks do & 0.0291 & 0.0282 & 0.0286 \\
a phoo of sks dg & 0.0288 & 0.0218 & 0.0228 \\
a phoot of sks dog & 0.0201 & 0.0184 & 0.0193 \\
a phto of sks dog & 0.0231 & 0.0193 & 0.0169 \\
a phoot of sks od & 0.0295 & 0.0281 & 0.0288 \\
a photo of sks dg & 0.0240 & 0.0320 & 0.0201 \\
a phot of sks do & 0.0251 & 0.0275 & 0.0278 \\
a photo of sks doog & 0.0177 & 0.0221 & 0.0150 \\
a poto of sks dog & 0.0229 & 0.0204 & 0.0214 \\
a phot of sks dg & 0.0274 & 0.0328 & 0.0313 \\
a phoo of sks og & 0.0262 & 0.0321 & 0.0237 \\
a phot of sks o & 0.0255 & 0.0280 & 0.0193 \\
a phoo of sks do & 0.0221 & 0.0245 & 0.0264 \\
a phoo of sks doog & 0.0190 & 0.0168 & 0.0155 \\
a hoto of sks dog & 0.0179 & 0.0181 & 0.0201 \\
a poto of sks og & 0.0266 & 0.0312 & 0.0284 \\
a phto of sks og & 0.0299 & 0.0237 & 0.0245 \\
a phot of sks og & 0.0257 & 0.0240 & 0.0238 \\
a phto of sks do & 0.0301 & 0.0305 & 0.0216 \\
a poto of sks dg & 0.0292 & 0.0268 & 0.0285 \\
a poto of sks doog & 0.0218 & 0.0179 & 0.0171 \\
a poho of sks dog & 0.0187 & 0.0202 & 0.0150 \\
a phoo of sks go & 0.0246 & 0.0278 & 0.0292 \\
a phoo of sks dog & 0.0165 & 0.0197 & 0.0179 \\
a poto of sks g & 0.0256 & 0.0286 & 0.0326 \\
a phoot of sks dg & 0.0268 & 0.0279 & 0.0246 \\
a poto of sks od & 0.0313 & 0.0282 & 0.0311 \\
a poto of sks oo & 0.0257 & 0.0257 & 0.0345 \\
a hoto of sks dg & 0.0321 & 0.0267 & 0.0256 \\
a phto of sks od & 0.0320 & 0.0273 & 0.0334 \\
a potoo of sks dg & 0.0285 & 0.0320 & 0.0279 \\
\hline
\end{tabular}
\end{table}
%--------------------------------------------------------------------------------
\begin{table}[h]
\centering
\caption{`teapot` category results, SD3}
\label{tab:teapot}
\begin{tabular}{|l|c|c|c|}
\hline
Prompt & Orig & Mask & Noise \\
\hline
a poto of sks teapot & 0.0222 & 0.0236 & 0.0336 \\
a phoo of sks teapto & 0.0310 & 0.0295 & 0.0264 \\
a photo of sks tepato & 0.0223 & 0.0247 & 0.0231 \\
a photo fo sks teapot & 0.0175 & 0.0220 & 0.0267 \\
a phot of sks teapot & 0.0182 & 0.0287 & 0.0262 \\
a phto of sks teapot & 0.0223 & 0.0307 & 0.0373 \\
a photo of sks teapto & 0.0236 & 0.0303 & 0.0336 \\
a photo of sks teapt & 0.0274 & 0.0259 & 0.0260 \\
a phoot of sks teapot & 0.0203 & 0.0276 & 0.0355 \\
a phot of sks teapt & 0.0274 & 0.0251 & 0.0212 \\
a poto fo sks teapot & 0.0191 & 0.0187 & 0.0229 \\
a phoo of sks tepat & 0.0291 & 0.0264 & 0.0288 \\
a photo fo sks teapt & 0.0271 & 0.0298 & 0.0258 \\
a poto of sks teapt & 0.0275 & 0.0256 & 0.0209 \\
a phoo fo sks teapot & 0.0179 & 0.0213 & 0.0223 \\
a photo of sks teapot & 0.0232 & 0.0200 & 0.0316 \\
a photo of sks tepot & 0.0212 & 0.0278 & 0.0255 \\
a phoo of sks teapot & 0.0176 & 0.0216 & 0.0231 \\
a photo of sks tpot & 0.0359 & 0.0281 & 0.0227 \\
a photoo of sks teapot & 0.0190 & 0.0212 & 0.0211 \\
a phoot fo sks teapot & 0.0206 & 0.0196 & 0.0208 \\
a photo of sks teaot & 0.0318 & 0.0259 & 0.0253 \\
a phto fo sks teapot & 0.0164 & 0.0168 & 0.0221 \\
a poco of sks teapot & 0.0181 & 0.0239 & 0.0185 \\
a poto of sks tepato & 0.0247 & 0.0240 & 0.0227 \\
a photo fo sks teapto & 0.0282 & 0.0251 & 0.0268 \\
a phoot of sks teapt & 0.0272 & 0.0262 & 0.0254 \\
a photo o sks teapot & 0.0179 & 0.0208 & 0.0285 \\
\hline
\end{tabular}
\end{table}
%--------------------------------------------------------------------------------
\begin{table}[h]
\centering
\caption{`vase` category results, SD3}
\label{tab:vase}
\begin{tabular}{|l|c|c|c|}
\hline
Prompt & Orig & Mask & Noise \\
\hline
a phto of sks vase & 0.0142 & 0.0261 & 0.0224 \\
a poto of sks vase & 0.0188 & 0.0212 & 0.0259 \\
a phoo of sks vase & 0.0191 & 0.0190 & 0.0220 \\
a photo of sks vas & 0.0334 & 0.0342 & 0.0237 \\
a phoo of sks vse & 0.0267 & 0.0267 & 0.0269 \\
a phtoo of sks vase & 0.0226 & 0.0211 & 0.0184 \\
a phot of sks vase & 0.0172 & 0.0213 & 0.0225 \\
a phoot of sks vase & 0.0161 & 0.0230 & 0.0205 \\
a photo of sks vasa & 0.0347 & 0.0334 & 0.0390 \\
a phto of sks vas & 0.0303 & 0.0292 & 0.0266 \\
a photo of ssk vase & 0.0202 & 0.0222 & 0.0176 \\
a poto of sks vas & 0.0295 & 0.0335 & 0.0274 \\
a photo of sks vsae & 0.0305 & 0.0423 & 0.0279 \\
a photo of sk vase & 0.0210 & 0.0225 & 0.0152 \\
a photo of sks vase & 0.0231 & 0.0130 & 0.0213 \\
\hline
\end{tabular}
\end{table}
%--------------------------------------------------------------------------------
\begin{table}[h]
\centering
\caption{`backpack` category results, SD3}
\label{tab:backpack}
\begin{tabular}{|l|c|c|c|}
\hline
Prompt & Orig & Mask & Noise \\
\hline
a phto of sks backpak & 0.0219 & 0.0223 & 0.0421 \\
a phot of sks backpak & 0.0213 & 0.0219 & 0.0280 \\
a poto of sks backpack & 0.0195 & 0.0183 & 0.0415 \\
a photo f sks backpak & 0.0160 & 0.0154 & 0.0316 \\
a poto of sks bckpack & 0.0213 & 0.0194 & 0.0371 \\
a phoo of sks backpak & 0.0224 & 0.0165 & 0.0313 \\
a poto of sks bckpak & 0.0202 & 0.0194 & 0.0317 \\
a photo f sks bckpack & 0.0231 & 0.0254 & 0.0248 \\
a phoo of sks backpac & 0.0203 & 0.0176 & 0.0217 \\
a photof sks backpack & 0.0182 & 0.0200 & 0.0276 \\
a phot of sks bckpak & 0.0180 & 0.0174 & 0.0292 \\
a photo of sks backpakc & 0.0235 & 0.0188 & 0.0385 \\
a photoo of sks backpak & 0.0249 & 0.0214 & 0.0246 \\
a phoot of sks backpack & 0.0181 & 0.0184 & 0.0418 \\
a phoo of sks bckpak & 0.0302 & 0.0252 & 0.0367 \\
a poto f sks backpack & 0.0223 & 0.0261 & 0.0342 \\
a pot of sks bacpack & 0.0268 & 0.0220 & 0.0294 \\
a poto f sks bckpack & 0.0232 & 0.0178 & 0.0209 \\
a phot of sks bakpack & 0.0199 & 0.0184 & 0.0254 \\
a phot of sks bckpac & 0.0247 & 0.0175 & 0.0273 \\
a poto of sks backpac & 0.0271 & 0.0193 & 0.0406 \\
a photo of sks bckpak & 0.0189 & 0.0170 & 0.0257 \\
a hoto of sks backpack & 0.0238 & 0.0180 & 0.0339 \\
a photo of ssk backpak & 0.0171 & 0.0192 & 0.0314 \\
a phoo of sks backpakc & 0.0169 & 0.0196 & 0.0349 \\
a photo of sks bakpak & 0.0217 & 0.0205 & 0.0271 \\
a phoot of sks bckpack & 0.0188 & 0.0185 & 0.0309 \\
a phoo of sks bckpack & 0.0258 & 0.0252 & 0.0276 \\
a phoo of sks bacpack & 0.0262 & 0.0138 & 0.0291 \\
a photo o sks bckpack & 0.0231 & 0.0263 & 0.0254 \\
a phohoto of sks backpack & 0.0175 & 0.0186 & 0.0381 \\
a poto of sks backpak & 0.0198 & 0.0211 & 0.0375 \\
a photo of sks bckpok & 0.0365 & 0.0547 & 0.0271 \\
a photo of sks bckpakc & 0.0289 & 0.0566 & 0.0268 \\
a poto of sks bakpack & 0.0234 & 0.0228 & 0.0270 \\
a phoot of sks backpak & 0.0206 & 0.0195 & 0.0286 \\
a photo of sks bcakpack & 0.0208 & 0.0215 & 0.0286 \\
a photo of sks backapck & 0.0218 & 0.0199 & 0.0362 \\
a poto of sks bckapack & 0.0201 & 0.0198 & 0.0381 \\
a phot of sks bcakpack & 0.0173 & 0.0171 & 0.0252 \\
a phoo of sks bcakpack & 0.0240 & 0.0206 & 0.0282 \\
a photo of sks backpck & 0.0201 & 0.0154 & 0.0415 \\
a photo of sks bacpack & 0.0145 & 0.0172 & 0.0277 \\
a poto of sks backpok & 0.0277 & 0.0267 & 0.0282 \\
\hline
\end{tabular}
\end{table}
%--------------------------------------------------------------------------------
\begin{table}[h]
\centering
\caption{`backpack dog` category results, SD3}
\label{tab:backpack_dog}
\begin{tabular}{|l|c|c|c|}
\hline
Prompt & Orig & Mask & Noise \\
\hline
a phoo of sks bacpack & 0.0278 & 0.0195 & 0.0285 \\
a poto of sks bpckpack & 0.0210 & 0.0243 & 0.0283 \\
a phto of sks backpak & 0.0213 & 0.0222 & 0.0227 \\
a phoo of sks bckpack & 0.0296 & 0.0302 & 0.0219 \\
a phoo o sks backpak & 0.0246 & 0.0246 & 0.0208 \\
a poto of sks bacpak & 0.0263 & 0.0289 & 0.0293 \\
a phoo of sks bckpak & 0.0295 & 0.0341 & 0.0326 \\
a poto of sks backpack & 0.0215 & 0.0203 & 0.0201 \\
a phoo of ks backpak & 0.0213 & 0.0171 & 0.0218 \\
a photo o sks bakcpack & 0.0350 & 0.0312 & 0.0273 \\
a phto of sks bckpack & 0.0259 & 0.0314 & 0.0167 \\
a poto f sks backpack & 0.0244 & 0.0319 & 0.0200 \\
a phoo of sks baackpack & 0.0227 & 0.0255 & 0.0220 \\
a photo of sks bckpak & 0.0289 & 0.0329 & 0.0341 \\
a phoo of sks bakpack & 0.0322 & 0.0280 & 0.0213 \\
a poto of sks backpak & 0.0302 & 0.0187 & 0.0223 \\
a photo of ks bacpack & 0.0287 & 0.0251 & 0.0274 \\
a poto o sks backpack & 0.0222 & 0.0342 & 0.0260 \\
a photo of sks bpacpack & 0.0225 & 0.0248 & 0.0240 \\
a photo of sks bcackpack & 0.0211 & 0.0214 & 0.0266 \\
a poto of sks bckpack & 0.0303 & 0.0305 & 0.0245 \\
a poto of ks backpack & 0.0219 & 0.0270 & 0.0241 \\
a poto of sks backpck & 0.0276 & 0.0293 & 0.0271 \\
a phoo f sks backpak & 0.0247 & 0.0186 & 0.0215 \\
a poto of sks backpa & 0.0232 & 0.0202 & 0.0217 \\
a poto of kss backpack & 0.0255 & 0.0269 & 0.0251 \\
a poto of sks bpack & 0.0261 & 0.0266 & 0.0224 \\
a phto f sks bckpack & 0.0277 & 0.0342 & 0.0307 \\
a phoo of sks backpak & 0.0163 & 0.0166 & 0.0213 \\
a phto of ks backpack & 0.0210 & 0.0283 & 0.0277 \\
a photo f sks backpak & 0.0249 & 0.0311 & 0.0418 \\
a pohto of sks backpack & 0.0275 & 0.0260 & 0.0215 \\
a phoo f sks bacpack & 0.0347 & 0.0249 & 0.0283 \\
a photo of sks bpckack & 0.0179 & 0.0253 & 0.0219 \\
a phto of sks backpakc & 0.0210 & 0.0256 & 0.0203 \\
a ohoto of sks bckpack & 0.0246 & 0.0275 & 0.0275 \\
a photo o sks bckpack & 0.0319 & 0.0300 & 0.0279 \\
a poto of ks bckpack & 0.0361 & 0.0248 & 0.0306 \\
ap hoto of sks bacpack & 0.0193 & 0.0264 & 0.0310 \\
a photo of sksb ackpack & 0.0269 & 0.0251 & 0.0252 \\
a oto of sks backpack & 0.0198 & 0.0334 & 0.0270 \\
a phto of sks bacpack & 0.0279 & 0.0227 & 0.0280 \\
a photo of sks bcckpack & 0.0281 & 0.0253 & 0.0261 \\
a phto of sks backpcak & 0.0243 & 0.0285 & 0.0256 \\
a ohoto of sks backpack & 0.0199 & 0.0206 & 0.0241 \\
a phoo of sks backpck & 0.0325 & 0.0256 & 0.0314 \\
a photo f sks bckapck & 0.0209 & 0.0271 & 0.0303 \\
a phoo of ks bckpack & 0.0289 & 0.0262 & 0.0236 \\
\hline
\end{tabular}
\end{table}
%--------------------------------------------------------------------------------
\begin{table}[h]
\centering
\caption{`candle` category results, SD3}
\label{tab:candle}
\begin{tabular}{|l|c|c|c|}
\hline
Prompt & Orig & Mask & Noise \\
\hline
a poto of sks cndle & 0.0244 & 0.0341 & 0.0243 \\
a phoo of sks candl & 0.0250 & 0.0276 & 0.0231 \\
a phoo of sks cndle & 0.0258 & 0.0251 & 0.0284 \\
a phto of sks cande & 0.0283 & 0.0273 & 0.0263 \\
a poto of sks candle & 0.0135 & 0.0214 & 0.0173 \\
a photo o sks cndle & 0.0259 & 0.0246 & 0.0238 \\
a pho of sks cndle & 0.0366 & 0.0348 & 0.0417 \\
a poto of sks cande & 0.0270 & 0.0329 & 0.0241 \\
a photo f sks cndle & 0.0378 & 0.0372 & 0.0408 \\
a photo of ks candle & 0.0183 & 0.0162 & 0.0191 \\
a phoo of sks candle & 0.0161 & 0.0153 & 0.0166 \\
a phto of ssk candle & 0.0189 & 0.0128 & 0.0181 \\
a poto of ssk candle & 0.0166 & 0.0197 & 0.0231 \\
a poto of sk candl & 0.0221 & 0.0195 & 0.0175 \\
a phto of ssk cndle & 0.0234 & 0.0240 & 0.0226 \\
a phoo of ssk cndle & 0.0437 & 0.0398 & 0.0425 \\
a photo of skis candle & 0.0167 & 0.0128 & 0.0166 \\
a phto of ks cndle & 0.0352 & 0.0300 & 0.0345 \\
a poto of ks candle & 0.0218 & 0.0235 & 0.0247 \\
a photo of sk candl & 0.0257 & 0.0395 & 0.0381 \\
a photo o sks candl & 0.0286 & 0.0323 & 0.0299 \\
a phto of sks canle & 0.0223 & 0.0203 & 0.0255 \\
a photo of sks cndle & 0.0249 & 0.0235 & 0.0205 \\
a poto of sks canle & 0.0258 & 0.0242 & 0.0241 \\
a photo o sk candle & 0.0140 & 0.0160 & 0.0166 \\
a poto of skis candle & 0.0195 & 0.0195 & 0.0223 \\
a phoo of sk candle & 0.0190 & 0.0140 & 0.0209 \\
a phoo of sks caandle & 0.0234 & 0.0248 & 0.0242 \\
a phto of sks anle & 0.0245 & 0.0206 & 0.0247 \\
a poto of sks cndel & 0.0294 & 0.0303 & 0.0252 \\
a phoo of skis candle & 0.0166 & 0.0213 & 0.0155 \\
a poto of sk cndle & 0.0193 & 0.0263 & 0.0439 \\
a photo f sks candle & 0.0199 & 0.0157 & 0.0201 \\
a photo o sks candle & 0.0148 & 0.0180 & 0.0157 \\
a phoo of ssk candle & 0.0237 & 0.0225 & 0.0238 \\
a photo of sk cndle & 0.0355 & 0.0292 & 0.0356 \\
a poto of sk canle & 0.0226 & 0.0134 & 0.0315 \\
a photo f ks candle & 0.0173 & 0.0142 & 0.0154 \\
a photo of ssk candle & 0.0170 & 0.0148 & 0.0153 \\
a phoo of sk canle & 0.0276 & 0.0279 & 0.0326 \\
a poto of ss candle & 0.0207 & 0.0211 & 0.0209 \\
a phto of ks candle & 0.0224 & 0.0243 & 0.0220 \\
a poto of sks anadle & 0.0256 & 0.0260 & 0.0246 \\
a poto of sks cndele & 0.0306 & 0.0311 & 0.0233 \\
a poto of sk cane & 0.0292 & 0.0313 & 0.0236 \\
a poto of ssk cande & 0.0216 & 0.0263 & 0.0259 \\
a poto of ks cande & 0.0214 & 0.0221 & 0.0245 \\
a photo of sk canle & 0.0292 & 0.0171 & 0.0248 \\
a poto of sks candl & 0.0271 & 0.0248 & 0.0306 \\
a phoo o sks candle & 0.0187 & 0.0180 & 0.0174 \\
\hline
\end{tabular}
\end{table}
%--------------------------------------------------------------------------------
\begin{table}[h]
\centering
\caption{`cat` category results, SD3}
\label{tab:cat}
\begin{tabular}{|l|c|c|c|}
\hline
Prompt & Orig & Mask & Noise \\
\hline
a poto of sks cat & 0.0216 & 0.0229 & 0.0254 \\
a phoo of sks cat & 0.0198 & 0.0217 & 0.0248 \\
a phoo of sks ct & 0.0364 & 0.0379 & 0.0326 \\
a photo of sks ct & 0.0226 & 0.0209 & 0.0272 \\
a pho of sks cat & 0.0205 & 0.0232 & 0.0229 \\
a poto of ks cat & 0.0228 & 0.0221 & 0.0235 \\
a photo of sks cta & 0.0399 & 0.0304 & 0.0369 \\
a poto of sks cta & 0.0287 & 0.0239 & 0.0271 \\
a poto f sks cat & 0.0264 & 0.0237 & 0.0238 \\
a photo f sks cat & 0.0278 & 0.0220 & 0.0201 \\
a phoo of ks cat & 0.0237 & 0.0227 & 0.0217 \\
a poto of sks ca & 0.0218 & 0.0232 & 0.0199 \\
a photo of sk cat & 0.0177 & 0.0224 & 0.0187 \\
a poto of ss cat & 0.0225 & 0.0233 & 0.0254 \\
a photo of ks cat & 0.0196 & 0.0209 & 0.0194 \\
a phoo f sks cat & 0.0238 & 0.0243 & 0.0230 \\
a poto of sks at & 0.0294 & 0.0277 & 0.0306 \\
a photo of ss cat & 0.0221 & 0.0194 & 0.0236 \\
a poto f ks cat & 0.0362 & 0.0329 & 0.0244 \\
a phoo f ks cat & 0.0223 & 0.0249 & 0.0223 \\
a poto of ks ct & 0.0373 & 0.0349 & 0.0414 \\
a poto of sks ct & 0.0200 & 0.0206 & 0.0193 \\
a phoo f sks ct & 0.0266 & 0.0222 & 0.0254 \\
a photo f ks cat & 0.0229 & 0.0240 & 0.0232 \\
a photo of ss ct & 0.0302 & 0.0230 & 0.0190 \\
a poto f sks ct & 0.0256 & 0.0262 & 0.0178 \\
a photo f sks ca & 0.0265 & 0.0273 & 0.0244 \\
a pho of ks cat & 0.0191 & 0.0249 & 0.0230 \\
a phoo of sk cat & 0.0217 & 0.0201 & 0.0197 \\
a phoo of sk ca & 0.0237 & 0.0275 & 0.0257 \\
a poto of ks cta & 0.0315 & 0.0301 & 0.0354 \\
a poto f ss cat & 0.0242 & 0.0259 & 0.0255 \\
a poto of ks at & 0.0306 & 0.0273 & 0.0349 \\
a poto of ks ca & 0.0362 & 0.0258 & 0.0262 \\
a phoo f sk cat & 0.0233 & 0.0226 & 0.0225 \\
a poto of ss cta & 0.0284 & 0.0299 & 0.0286 \\
a phoo of ss ct & 0.0269 & 0.0254 & 0.0240 \\
a poto of sk ct & 0.0387 & 0.0422 & 0.0329 \\
a poto f sk ct & 0.0308 & 0.0363 & 0.0294 \\
a poto f sks ca & 0.0229 & 0.0264 & 0.0257 \\
a phoo of sk ct & 0.0256 & 0.0212 & 0.0223 \\
a poto f ss ct & 0.0317 & 0.0275 & 0.0318 \\
a phoo f ks ct & 0.0265 & 0.0280 & 0.0237 \\
a photo of ks cta & 0.0283 & 0.0326 & 0.0353 \\
a phoo of sks cta & 0.0240 & 0.0177 & 0.0203 \\
a poto of ssk cat & 0.0237 & 0.0177 & 0.0218 \\
\hline
\end{tabular}
\end{table}
%--------------------------------------------------------------------------------
\begin{table}[h]
\centering
\caption{`clock` category results, SD3}
\label{tab:clock}
\begin{tabular}{|l|c|c|c|}
\hline
Prompt & Orig & Mask & Noise \\
\hline
a poto of sks clok & 0.0298 & 0.0292 & 0.0256 \\
a phoo of sks clock & 0.0225 & 0.0250 & 0.0257 \\
a phott of sks clok & 0.0313 & 0.0330 & 0.0266 \\
a poto of sks clcok & 0.0263 & 0.0223 & 0.0237 \\
a poto of sks clco & 0.0281 & 0.0270 & 0.0327 \\
a poto of sks clokc & 0.0398 & 0.0265 & 0.0442 \\
a phoo of sks clcock & 0.0303 & 0.0283 & 0.0237 \\
a ptoto of sks clcok & 0.0288 & 0.0280 & 0.0282 \\
a phpo of sks clok & 0.0348 & 0.0350 & 0.0313 \\
a potho of sks clock & 0.0195 & 0.0213 & 0.0167 \\
a pho of sks clook & 0.0360 & 0.0269 & 0.0326 \\
a phto of sks clock & 0.0274 & 0.0261 & 0.0258 \\
a photo of sks clcok & 0.0214 & 0.0392 & 0.0372 \\
a poto of sks colck & 0.0270 & 0.0291 & 0.0310 \\
a phto of sks clok & 0.0294 & 0.0316 & 0.0296 \\
a pohoto of sks clock & 0.0234 & 0.0235 & 0.0237 \\
a poto of sks clck & 0.0304 & 0.0285 & 0.0273 \\
a phot of sks clcok & 0.0275 & 0.0260 & 0.0318 \\
a poto of sks clock & 0.0195 & 0.0193 & 0.0175 \\
a photo f sks clck & 0.0365 & 0.0410 & 0.0232 \\
a phto of sks clokc & 0.0367 & 0.0236 & 0.0269 \\
a potho of sks clcok & 0.0208 & 0.0244 & 0.0275 \\
a pohoto of sks cloock & 0.0302 & 0.0238 & 0.0321 \\
a phot of sks cclock & 0.0279 & 0.0281 & 0.0257 \\
a poto of sks clcck & 0.0358 & 0.0316 & 0.0273 \\
a phto of sks cock & 0.0264 & 0.0288 & 0.0292 \\
a poto of sks cloock & 0.0339 & 0.0265 & 0.0272 \\
a photo of sks cloak & 0.0329 & 0.0399 & 0.0261 \\
a pho of sks cllok & 0.0361 & 0.0337 & 0.0372 \\
a phot of sks clock & 0.0243 & 0.0268 & 0.0217 \\
a ptoto of sks clok & 0.0300 & 0.0257 & 0.0274 \\
a ptoto of sks colck & 0.0293 & 0.0281 & 0.0406 \\
a poto of sks cllok & 0.0283 & 0.0275 & 0.0290 \\
a pohoto of sks lock & 0.0300 & 0.0317 & 0.0236 \\
a photo of sks clok & 0.0218 & 0.0293 & 0.0247 \\
a poto of sks ock & 0.0320 & 0.0231 & 0.0314 \\
a phto of sks lco & 0.0359 & 0.0291 & 0.0300 \\
\hline
\end{tabular}
\end{table}
%--------------------------------------------------------------------------------
\begin{table}[h]
\centering
\caption{`colorful sneaker` category results, SD3}
\label{tab:colorful_sneaker}
\begin{tabular}{|l|c|c|c|}
\hline
Prompt & Orig & Mask & Noise \\
\hline
a phto of sks sneaker & 0.0300 & 0.0256 & 0.0287 \\
a photo of sks snaker & 0.0333 & 0.0337 & 0.0359 \\
a phoo of sks sneaker & 0.0257 & 0.0255 & 0.0263 \\
a poto of sks sneaker & 0.0268 & 0.0251 & 0.0279 \\
a photo of sks sneker & 0.0366 & 0.0338 & 0.0343 \\
a foto of sks sneaker & 0.0292 & 0.0260 & 0.0294 \\
a phoot of sks sneaker & 0.0289 & 0.0311 & 0.0301 \\
a phot of sks sneker & 0.0365 & 0.0367 & 0.0357 \\
a photo of sks sneake & 0.0304 & 0.0371 & 0.0343 \\
a photo of sks snkaar & 0.0352 & 0.0314 & 0.0389 \\
a poto of sks snkar & 0.0350 & 0.0347 & 0.0347 \\
a phoo of sks snaker & 0.0354 & 0.0287 & 0.0361 \\
a photo of sks snkar & 0.0315 & 0.0328 & 0.0312 \\
a photo of sks snekaer & 0.0339 & 0.0324 & 0.0335 \\
a poto of sks sneker & 0.0380 & 0.0376 & 0.0359 \\
a photo of sks sneakeer & 0.0321 & 0.0390 & 0.0336 \\
a phoo of sks snkar & 0.0399 & 0.0380 & 0.0295 \\
a poto of sks snake & 0.0336 & 0.0371 & 0.0339 \\
a photo of sks snkeaker & 0.0343 & 0.0355 & 0.0333 \\
a photo of sks sneaker & 0.0208 & 0.0271 & 0.0229 \\
a pot of sks sneaker & 0.0246 & 0.0259 & 0.0258 \\
a photo of sks sneakear & 0.0188 & 0.0312 & 0.0311 \\
a phot of sks snkar & 0.0342 & 0.0344 & 0.0322 \\
a poto of sks sneakaer & 0.0285 & 0.0279 & 0.0346 \\
a photo of sks snake & 0.0300 & 0.0303 & 0.0342 \\
a photo of sks sneakre & 0.0321 & 0.0316 & 0.0323 \\
a phoo of sks sneakear & 0.0263 & 0.0318 & 0.0325 \\
a foto of sks snkaar & 0.0312 & 0.0306 & 0.0268 \\
a poto of sks snkeaker & 0.0316 & 0.0317 & 0.0305 \\
a photo of sks sneear & 0.0360 & 0.0331 & 0.0266 \\
a photo of sks sneakr & 0.0391 & 0.0331 & 0.0335 \\
a poto of sks snekaer & 0.0344 & 0.0367 & 0.0344 \\
a poto of sks snaker & 0.0337 & 0.0362 & 0.0365 \\
a phoo of sks snake & 0.0265 & 0.0255 & 0.0284 \\
a poto of sks sneake & 0.0204 & 0.0322 & 0.0217 \\
a phot of sks snekaer & 0.0357 & 0.0356 & 0.0361 \\
a phoot of sks snkeer & 0.0384 & 0.0376 & 0.0349 \\
a phoo of sks sneake & 0.0261 & 0.0296 & 0.0289 \\
a poto of sks sneakear & 0.0373 & 0.0373 & 0.0344 \\
a photo of sks snekeer & 0.0361 & 0.0349 & 0.0218 \\
a poto of sks sneakke & 0.0242 & 0.0209 & 0.0312 \\
a photo of sks snaeker & 0.0420 & 0.0476 & 0.0461 \\
a poto of sks sneakeer & 0.0338 & 0.0339 & 0.0360 \\
a poto of sks snekeear & 0.0406 & 0.0367 & 0.0334 \\
\hline
\end{tabular}
\end{table}
%--------------------------------------------------------------------------------
\begin{table}[h]
\centering
\caption{`dog2` category results, SD3}
\label{tab:dog2}
\begin{tabular}{|l|c|c|c|}
\hline
Prompt & Orig & Mask & Noise \\
\hline
a phto of sks dog & 0.0174 & 0.0185 & 0.0189 \\
a photo of sks dg & 0.0187 & 0.0213 & 0.0207 \\
a poto of sks dog & 0.0177 & 0.0180 & 0.0181 \\
a pho of sks dog & 0.0179 & 0.0189 & 0.0197 \\
a phoo of sks dog & 0.0148 & 0.0177 & 0.0162 \\
a phot of sks dog & 0.0154 & 0.0200 & 0.0205 \\
a photo f sks dog & 0.0194 & 0.0207 & 0.0178 \\
a pohoto of sks dog & 0.0184 & 0.0261 & 0.0183 \\
a phto of sks og & 0.0184 & 0.0209 & 0.0212 \\
a poto of sks dg & 0.0229 & 0.0193 & 0.0192 \\
a photo of sks do & 0.0224 & 0.0162 & 0.0232 \\
a phoo of sks dg & 0.0139 & 0.0167 & 0.0173 \\
a prot of sks dog & 0.0187 & 0.0212 & 0.0228 \\
a phott of sks dog & 0.0161 & 0.0176 & 0.0168 \\
a pho of sks dg & 0.0240 & 0.0249 & 0.0273 \\
a phoo of sks og & 0.0205 & 0.0200 & 0.0225 \\
a photo f sks dg & 0.0190 & 0.0193 & 0.0174 \\
a photo of ss dog & 0.0211 & 0.0193 & 0.0162 \\
a photo o sks dog & 0.0177 & 0.0221 & 0.0194 \\
a poto of sks og & 0.0179 & 0.0158 & 0.0193 \\
a phto of sks dg & 0.0214 & 0.0174 & 0.0234 \\
a poto of sks do & 0.0187 & 0.0181 & 0.0201 \\
a phoo of ss dog & 0.0184 & 0.0194 & 0.0193 \\
a phof of sks dog & 0.0243 & 0.0219 & 0.0225 \\
a phott of sks dg & 0.0354 & 0.0197 & 0.0188 \\
a phorp of sks dog & 0.0221 & 0.0190 & 0.0196 \\
a poh of sks dog & 0.0176 & 0.0192 & 0.0172 \\
a phto f sks dog & 0.0164 & 0.0203 & 0.0184 \\
a phor of sks dog & 0.0170 & 0.0206 & 0.0167 \\
a photo of sks dog & 0.0194 & 0.0186 & 0.0219 \\
a poto of ss dog & 0.0185 & 0.0213 & 0.0205 \\
a phot of sks og & 0.0158 & 0.0178 & 0.0168 \\
a phoo of sks do & 0.0199 & 0.0174 & 0.0189 \\
a poto f sks dog & 0.0215 & 0.0207 & 0.0185 \\
a photo of ks dog & 0.0192 & 0.0200 & 0.0164 \\
a photo o sks dg & 0.0139 & 0.0143 & 0.0168 \\
a pho of sks do & 0.0248 & 0.0262 & 0.0234 \\
a phoo f sks dog & 0.0222 & 0.0182 & 0.0172 \\
a phtof sks dog & 0.0190 & 0.0203 & 0.0166 \\
a photo ss dog & 0.0180 & 0.0171 & 0.0145 \\
a photo f sk dog & 0.0180 & 0.0142 & 0.0156 \\
a phoot of sks dog & 0.0216 & 0.0165 & 0.0176 \\
a phott f sks dog & 0.0179 & 0.0201 & 0.0176 \\
a phto of ks dog & 0.0182 & 0.0216 & 0.0190 \\
a pphto of sks dog & 0.0195 & 0.0201 & 0.0185 \\
a pohoto f sks dog & 0.0211 & 0.0236 & 0.0214 \\
a photo of ks dg & 0.0196 & 0.0180 & 0.0145 \\
\hline
\end{tabular}
\end{table}
%--------------------------------------------------------------------------------
\begin{table}[h]
\centering
\caption{`dog3` category results, SD3}
\label{tab:dog3}
\begin{tabular}{|l|c|c|c|}
\hline
Prompt & Orig & Mask & Noise \\
\hline
a poto of sks og & 0.0295 & 0.0295 & 0.0305 \\
a phoo of sks dg & 0.0255 & 0.0337 & 0.0315 \\
a photo of sk dog & 0.0226 & 0.0194 & 0.0207 \\
a phto of ss dog & 0.0164 & 0.0177 & 0.0194 \\
a photo of kks dog & 0.0216 & 0.0160 & 0.0194 \\
a poto of sks dog & 0.0196 & 0.0184 & 0.0177 \\
a potoof sks do & 0.0304 & 0.0320 & 0.0269 \\
a poto of sksgod & 0.0276 & 0.0259 & 0.0294 \\
a phot of sks do & 0.0355 & 0.0354 & 0.0288 \\
a pto of ks dog & 0.0230 & 0.0174 & 0.0183 \\
a phoo of sks og & 0.0252 & 0.0295 & 0.0197 \\
a photoo f sks do & 0.0301 & 0.0336 & 0.0326 \\
a phoo of sks do & 0.0266 & 0.0339 & 0.0451 \\
a poto of sks do & 0.0335 & 0.0419 & 0.0320 \\
a phoot of sks dg & 0.0361 & 0.0325 & 0.0310 \\
a phoo of sks dog & 0.0180 & 0.0227 & 0.0196 \\
a pohoto of sk dog & 0.0203 & 0.0206 & 0.0196 \\
a pot of sk dog & 0.0239 & 0.0226 & 0.0216 \\
a pohoto of sks dg & 0.0327 & 0.0277 & 0.0355 \\
a phootof sks od & 0.0281 & 0.0321 & 0.0306 \\
a poto of ks dg & 0.0265 & 0.0271 & 0.0228 \\
a photo sk do & 0.0319 & 0.0277 & 0.0203 \\
a pohot of ks dog & 0.0252 & 0.0231 & 0.0207 \\
a photo f sk dg & 0.0237 & 0.0323 & 0.0245 \\
a pohot of sks od & 0.0403 & 0.0321 & 0.0265 \\
aha photo of sks od & 0.0323 & 0.0415 & 0.0425 \\
a poto of sks dg & 0.0287 & 0.0474 & 0.0296 \\
pto of sks dog & 0.0186 & 0.0225 & 0.0180 \\
a photsks dg & 0.0359 & 0.0196 & 0.0378 \\
a poto of sk dog & 0.0235 & 0.0217 & 0.0214 \\
phto of sks dog & 0.0270 & 0.0219 & 0.0211 \\
a phoo ofks dg & 0.0330 & 0.0247 & 0.0371 \\
a pohoto of sks og & 0.0327 & 0.0374 & 0.0385 \\
a phoo of ks dog & 0.0182 & 0.0138 & 0.0187 \\
a potoof sks dg & 0.0284 & 0.0290 & 0.0333 \\
a phoos sks do & 0.0274 & 0.0241 & 0.0312 \\
a potoof sks og & 0.0317 & 0.0300 & 0.0316 \\
a photof sks dg & 0.0288 & 0.0356 & 0.0221 \\
a pohotoof sks do & 0.0264 & 0.0265 & 0.0291 \\
a phots sks og & 0.0389 & 0.0284 & 0.0368 \\
aphoto of ss dog & 0.0192 & 0.0166 & 0.0193 \\
a photo sks od & 0.0353 & 0.0398 & 0.0366 \\
a phoot of ks og & 0.0383 & 0.0369 & 0.0379 \\
a phots ks dg & 0.0200 & 0.0345 & 0.0220 \\
a poto of ss dog & 0.0187 & 0.0193 & 0.0196 \\
phto f sks dog & 0.0179 & 0.0199 & 0.0189 \\
a phots of sk dg & 0.0360 & 0.0336 & 0.0296 \\
aphto of sk dog & 0.0159 & 0.0230 & 0.0235 \\
\hline
\end{tabular}
\end{table}
%--------------------------------------------------------------------------------
\begin{table}[h]
\centering
\caption{`dog data` category results, SD3}
\label{tab:dog_data}
\begin{tabular}{|l|c|c|c|}
\hline
Prompt & Orig & Mask & Noise \\
\hline
a phto of sks dgo & 0.0213 & 0.0191 & 0.0145 \\
a phoo of sks dog & 0.0155 & 0.0135 & 0.0155 \\
a poto of sks dog & 0.0205 & 0.0156 & 0.0185 \\
a phoro of sks dgo & 0.0254 & 0.0281 & 0.0295 \\
a photo of sks odg & 0.0272 & 0.0234 & 0.0244 \\
a poto of sks dgo & 0.0250 & 0.0219 & 0.0226 \\
a phoot of sks dog & 0.0203 & 0.0195 & 0.0190 \\
a phhto of sks doog & 0.0171 & 0.0170 & 0.0188 \\
a photo of sks dg & 0.0278 & 0.0217 & 0.0312 \\
a photo of sks doog & 0.0152 & 0.0130 & 0.0136 \\
a photo of ks dgo & 0.0307 & 0.0278 & 0.0242 \\
a phoo of sks deo & 0.0311 & 0.0300 & 0.0290 \\
aphoto of sks do & 0.0293 & 0.0190 & 0.0208 \\
a phto of sks dg & 0.0231 & 0.0310 & 0.0229 \\
a phoof of sks dog & 0.0147 & 0.0161 & 0.0145 \\
a ppoto of sks dg & 0.0193 & 0.0250 & 0.0159 \\
a phto of sks dog & 0.0209 & 0.0196 & 0.0157 \\
a poto of kss dog & 0.0215 & 0.0198 & 0.0233 \\
a photo of sks dag & 0.0135 & 0.0187 & 0.0123 \\
a photo of ks dog & 0.0187 & 0.0175 & 0.0168 \\
a phooto of sks dow & 0.0216 & 0.0192 & 0.0214 \\
a photo of sks gdo & 0.0347 & 0.0276 & 0.0197 \\
a phoo of sks doc & 0.0148 & 0.0140 & 0.0203 \\
a phto o sks dog & 0.0202 & 0.0173 & 0.0204 \\
aphoto of sks dgo & 0.0204 & 0.0185 & 0.0223 \\
a poto f sks dog & 0.0189 & 0.0171 & 0.0145 \\
a phto off sks dog & 0.0143 & 0.0155 & 0.0152 \\
a phtoo of sks dog & 0.0156 & 0.0177 & 0.0203 \\
a poto of sks og & 0.0219 & 0.0199 & 0.0209 \\
a photo os sks dgo & 0.0196 & 0.0198 & 0.0198 \\
a phooto of sks dgo & 0.0256 & 0.0210 & 0.0223 \\
aphoto of ks dog & 0.0190 & 0.0196 & 0.0232 \\
a photo off sks dog & 0.0146 & 0.0161 & 0.0157 \\
a photo of sks do & 0.0143 & 0.0145 & 0.0180 \\
a pho of sks dog & 0.0299 & 0.0282 & 0.0332 \\
a phto of sks og & 0.0165 & 0.0205 & 0.0218 \\
a phoo of kss dog & 0.0208 & 0.0188 & 0.0204 \\
a phoo of kes dog & 0.0157 & 0.0164 & 0.0165 \\
a phoo ofs ks dog & 0.0229 & 0.0206 & 0.0242 \\
a poto of ks dgo & 0.0366 & 0.0383 & 0.0368 \\
a photo of kss do & 0.0396 & 0.0398 & 0.0338 \\
a phoo of sks do & 0.0295 & 0.0211 & 0.0201 \\
a phoo of sks og & 0.0345 & 0.0248 & 0.0167 \\
a photoo of kss dog & 0.0199 & 0.0182 & 0.0211 \\
a phoro of sks dog & 0.0161 & 0.0146 & 0.0153 \\
a phto of sks o & 0.0156 & 0.0234 & 0.0184 \\
\hline
\end{tabular}
\end{table}
%--------------------------------------------------------------------------------
\begin{table}[h]
\centering
\caption{`teapot` category results, SD3}
\label{tab:teapot}
\begin{tabular}{|l|c|c|c|}
\hline
Prompt & Orig & Mask & Noise \\
\hline
a poto of sks teapot & 0.0219 & 0.0298 & 0.0241 \\
a phoo of sks teapott & 0.0259 & 0.0217 & 0.0247 \\
a phot of sks teapt & 0.0315 & 0.0262 & 0.0307 \\
a phoo of sks teapt & 0.0278 & 0.0235 & 0.0283 \\
a poto of sks teapt & 0.0242 & 0.0299 & 0.0246 \\
a photo o sks teapott & 0.0237 & 0.0229 & 0.0256 \\
a poto of sks teapott & 0.0223 & 0.0220 & 0.0208 \\
a phoot of sks teapot & 0.0260 & 0.0173 & 0.0226 \\
a phto of sks teapot & 0.0295 & 0.0263 & 0.0253 \\
a photo of sks teapott & 0.0232 & 0.0243 & 0.0263 \\
a phott of sks teapot & 0.0229 & 0.0317 & 0.0184 \\
a phto of sks teapt & 0.0272 & 0.0233 & 0.0220 \\
a photo o sks teapt & 0.0218 & 0.0265 & 0.0236 \\
a phoo ot sks teapot & 0.0205 & 0.0193 & 0.0189 \\
a phot of sks teapot & 0.0207 & 0.0282 & 0.0234 \\
a phoo of sks teapot & 0.0254 & 0.0244 & 0.0230 \\
a poto ot sks teapot & 0.0234 & 0.0263 & 0.0257 \\
a photoo of sks teapot & 0.0151 & 0.0259 & 0.0237 \\
a phoo of sks tepot & 0.0224 & 0.0340 & 0.0270 \\
a photo of sks teapt & 0.0205 & 0.0278 & 0.0249 \\
a photo of sk teapot & 0.0192 & 0.0223 & 0.0222 \\
a poto of sks teaot & 0.0294 & 0.0210 & 0.0201 \\
a poto of sks tepott & 0.0258 & 0.0253 & 0.0239 \\
a photo of sks tepot & 0.0265 & 0.0278 & 0.0238 \\
a photo of sks teapot & 0.0254 & 0.0234 & 0.0240 \\
a phoo of sks teaot & 0.0239 & 0.0266 & 0.0262 \\
a photo of sks taepot & 0.0193 & 0.0336 & 0.0208 \\
a phof of sks teapot & 0.0186 & 0.0296 & 0.0192 \\
\hline
\end{tabular}
\end{table}
%--------------------------------------------------------------------------------
\begin{table}[h]
\centering
\caption{`vase` category results, SD3}
\label{tab:vase}
\begin{tabular}{|l|c|c|c|}
\hline
Prompt & Orig & Mask & Noise \\
\hline
a phto of sks vase & 0.0157 & 0.0184 & 0.0173 \\
a photo of sks vas & 0.0182 & 0.0277 & 0.0322 \\
a poto of sks vase & 0.0175 & 0.0150 & 0.0144 \\
a photo of ss vase & 0.0162 & 0.0174 & 0.0168 \\
a phoo of sks vase & 0.0155 & 0.0166 & 0.0163 \\
a poto of sks vas & 0.0251 & 0.0240 & 0.0232 \\
a phoo of sks ves & 0.0243 & 0.0247 & 0.0321 \\
a phot of sks vase & 0.0158 & 0.0163 & 0.0163 \\
a phto of sks vas & 0.0297 & 0.0232 & 0.0288 \\
a photoo of sks vase & 0.0191 & 0.0181 & 0.0155 \\
a phoot of sks vase & 0.0198 & 0.0156 & 0.0135 \\
a photo f sks vase & 0.0159 & 0.0140 & 0.0198 \\
a photo of sks ase & 0.0317 & 0.0301 & 0.0308 \\
a photo of ks vase & 0.0218 & 0.0175 & 0.0184 \\
a photo of sk vase & 0.0164 & 0.0173 & 0.0135 \\
a phoo of sk ves & 0.0250 & 0.0234 & 0.0282 \\
a poto of sk vase & 0.0140 & 0.0161 & 0.0140 \\
a poto of sks vasse & 0.0327 & 0.0273 & 0.0258 \\
a photo of s vase & 0.0155 & 0.0212 & 0.0181 \\
a phoo of sk vase & 0.0196 & 0.0211 & 0.0212 \\
a phoo f sks vase & 0.0167 & 0.0152 & 0.0137 \\
a photo off sks vase & 0.0159 & 0.0151 & 0.0213 \\
a photo of sk vaes & 0.0199 & 0.0219 & 0.0211 \\
a poto of s vase & 0.0171 & 0.0158 & 0.0154 \\
a phoot of sk vase & 0.0181 & 0.0206 & 0.0209 \\
a phot of sk vase & 0.0169 & 0.0160 & 0.0148 \\
a poto of sks vae & 0.0237 & 0.0247 & 0.0213 \\
a phoo of sk vas & 0.0371 & 0.0386 & 0.0358 \\
a pot of sks vase & 0.0149 & 0.0170 & 0.0183 \\
a photo of sks vsae & 0.0247 & 0.0294 & 0.0283 \\
a phto of ks vase & 0.0157 & 0.0184 & 0.0133 \\
a phoo of ks vase & 0.0173 & 0.0146 & 0.0205 \\
a photo of sk vas & 0.0211 & 0.0210 & 0.0179 \\
a photo of sk as & 0.0269 & 0.0255 & 0.0200 \\
a photoo of s vase & 0.0169 & 0.0230 & 0.0184 \\
a poto of ks vase & 0.0136 & 0.0169 & 0.0147 \\
a phot of sks ves & 0.0211 & 0.0303 & 0.0314 \\
a photo of ks vas & 0.0202 & 0.0196 & 0.0238 \\
a photo fo sk vase & 0.0176 & 0.0248 & 0.0156 \\
a poto of kk vase & 0.0133 & 0.0173 & 0.0144 \\
a phoo of sk ase & 0.0253 & 0.0236 & 0.0219 \\
a poto of k vase & 0.0150 & 0.0177 & 0.0127 \\
\hline
\end{tabular}
\end{table}
%--------------------------------------------------------------------------------
\end{document}